%% file: main.tex
\documentclass[runningheads]{llncs}

\usepackage{eccv}

\usepackage{eccvabbrv}

\usepackage{graphicx}
\usepackage{booktabs}
\usepackage{colortbl}
\usepackage{amssymb}
\usepackage{amsfonts}
\usepackage{amsmath}

\usepackage{booktabs}
\usepackage{enumitem}
\usepackage{microtype}
\usepackage[dvipsnames]{xcolor}
\usepackage{bm, bbm}
\usepackage{algorithmic}
\usepackage[linesnumbered,boxed,ruled,commentsnumbered]{algorithm2e}
\usepackage{multirow}

\usepackage{floatrow}
\newfloatcommand{capbtabbox}{table}[][\FBwidth]

\usepackage{marvosym}

\usepackage[accsupp]{axessibility}  

\newcommand{\hlr}[1]{{\color{red}{#1}}}
\newcommand{\hlb}[1]{{\color{blue}{#1}}}
\newcommand{\hlg}[1]{{\color{green}{#1}}}
\newcommand{\hly}[1]{{\color{yellow}{#1}}}

\usepackage{hyperref}

\usepackage{orcidlink}

\makeatletter
\def\thanks#1{\protected@xdef\@thanks{\@thanks
        \protect\footnotetext{#1}}}

\begin{document}

\title{SimPB: A Single Model for 2D and 3D Object Detection from Multiple Cameras} 

\titlerunning{SimPB}

\author{
Yingqi Tang\inst{\star} \thanks{
\inst{\star} Equal contribution.} \and
Zhaotie Meng\inst{\star} \and
Guoliang Chen \and
Erkang Cheng\textsuperscript{\Letter}
\thanks{
\textsuperscript{\Letter} Corresponding author.
}
}

\authorrunning{Y.~Tang et al.}


\institute{
Nullmax \\
\email{\{tangyingqi,mengzhaotie,chenguoliang,chengerkang\}@nullmax.ai}
}

\maketitle
\input{sec/0_abstract}    
\input{sec/1_intro}

\input{sec/2_relatedwork}
\input{sec/3_method}
\input{sec/4_experiments}

\input{sec/5_conclusion}

%
%

\bibliographystyle{splncs04}
\bibliography{main}

\clearpage
\input{sec/supplementary}

\end{document}

%% file: sec/0_abstract.tex
\begin{abstract}
The field of autonomous driving has attracted considerable interest in approaches that directly infer 3D objects in the Bird's Eye View (BEV) from multiple cameras. 
Some attempts have also explored utilizing 2D detectors from single images to enhance the performance of 3D detection.
However, these approaches rely on a two-stage process with separate detectors, where the 2D detection results are utilized only once for token selection or query initialization.
In this paper, we present a single model termed \textit{SimPB}, which \textbf{Sim}ultaneously detects 2D objects in the \textbf{P}erspective view and 3D objects in the \textbf{B}EV space from multiple cameras.
To achieve this, we introduce a hybrid decoder consisting of several multi-view 2D decoder layers and several 3D decoder layers, specifically designed for their respective detection tasks.
A Dynamic Query Allocation module and an Adaptive Query Aggregation module are proposed to continuously update and refine the interaction between 2D and 3D results, in a cyclic 3D-2D-3D manner.
Additionally, Query-group Attention is utilized to strengthen the interaction among 2D queries within each camera group. 
In the experiments, we evaluate our method on the nuScenes dataset and demonstrate promising results for both 2D and 3D detection tasks.
Our code is available at:~\url{https://github.com/nullmax-vision/SimPB}.
\keywords{Autonomous Driving \and 3D Object Detection \and Transformer}
\end{abstract}

%% file: sec/1_intro.tex
\begin{figure}[t]
  \centering
  \includegraphics[width=0.9\textwidth]{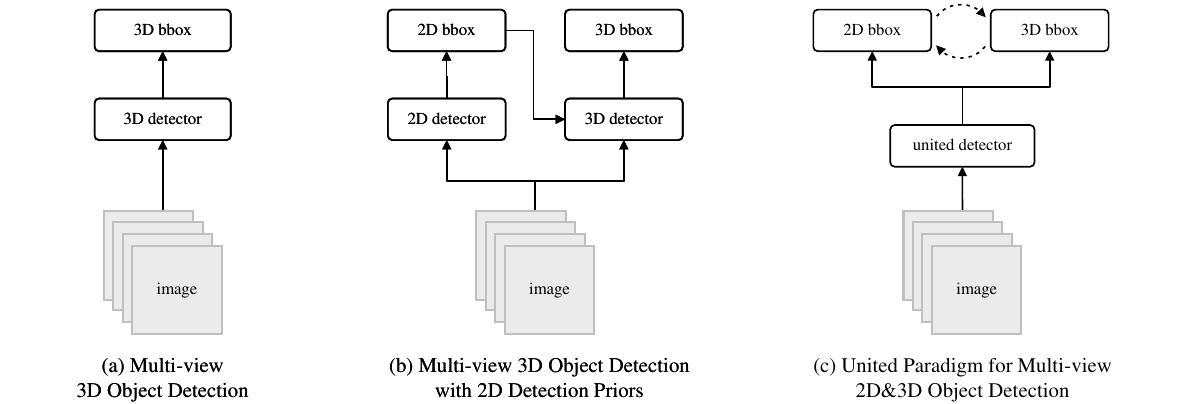}
  \caption{Comparisons of different multi-view object detection pipelines. (a) Multi-view 3D object detection. (b) A two-stage multi-view 3D object detector where 2D box detection is used as token selection or 3D query initialization. (c) Our proposed unified paradigm simultaneously predicts 2D and 3D results in a single model.}
  \label{fig:1_structure_compare}
\end{figure}

\section{Introduction}
\label{sec:intriduction}

The camera serves as a cost-effective sensor and plays a crucial role in the perception system of autonomous driving. 
In recent years, significant advancements have been made in 2D object detection for the perspective view~\cite{FasterRCNN, YoloV1, FCOS}. Building upon these methods, monocular 3D object detection techniques allow for the direct generation of 3D object results in the bird's eye view (BEV) space~\cite{Monocular3DOD, M3D-RPN}.
In contrast to using a single frontal-facing camera, multi-view 3D object detection \cite{FCOS3D, DETR3D, BEVFormer} utilizing several surrounding cameras has gained significant traction. It can provide comprehensive perception results for the ego vehicle, leveraging the combined viewpoints captured by multiple cameras.

To reduce the complex fusion step of monocular 3D detection from each surrounding camera, current multi-view 3D object detection directly computes the 3D object results (as in~\cref{fig:1_structure_compare} (a)). It can be roughly categorized into dense BEV methods and sparse query-based methods. 
Dense BEV methods~\cite{LSS, BEVFormer, BEVDepth, BEVNeXt} first transform image features from multiple cameras into a BEV feature map and then apply a 3D detector head to compute 3D objects.
Building a large BEV map for long-range objects poses challenges due to the high computational resources required for dense BEV representations.
Instead, sparse query-based methods~\cite{DETR3D, PETR, Sparse4D, StreamPETR, SparseBEV} employ a DETR-like scheme~\cite{DETR} to interact with feature maps from multiple cameras and eliminate the need for dense BEV feature construction.
These approaches typically employ pre-trained backbones from popular 2D perception tasks. While they deliver impressive performance, they do not fully leverage the benefits of off-the-shelf improvements in 2D detectors.

2D detection on the perspective view is valuable for multi-view autonomous driving perception systems. 
It offers an alternative approach to computing 3D object information by utilizing 2D boxes from multiple cameras in post-processing steps.  
In practical perception systems, this process is commonly implemented using two independent detectors.
Several query-based methods~\cite{FocalPETR, MV2D, Far3D} have been also developed to enhance the performance of 3D detection by incorporating 2D detection results of different cameras (as in~\cref{fig:1_structure_compare} (b)).
To generate 3D anchors, MV2D~\cite{MV2D} explicitly transfers the 2D box to a 2.5D point by leveraging the ROI image feature and camera parameters. 
Far3D~\cite{Far3D} employs a different approach by lifting the 2D box to 3D using depth estimation from a separate network.
Focal-PETR~\cite{FocalPETR} incorporates instance-guided supervision on perspective views to identify discriminative image tokens.
Using strong 2D detection results, these methods improve 3D object detection precision and recall by initializing 3D queries or providing selective foreground semantic information.

These two detector methods exhibit the following issues:
(1) The 2D detector treats the same object captured by different cameras as separate instances in the image space. As a result, the 3D detector tends to focus on local parts rather than capturing the global information of an object.
(2) The two-detector approach operates in a two-stage manner, where 2D information is only used once during token selection or query initialization. This lack of iterative updating of object semantics from 2D detection in the 3D detector leads to the underutilization of 2D features in 3D detection.
(3) Additionally, the use of different approaches for 2D and 3D detection, such as a CNN-based 2D detector and a Transformer-based 3D detector, introduces challenges in model optimization.

In this paper, we propose SimPB, a unified query-based detector that simultaneously detects 2D objects in the perspective view and 3D objects in the BEV space from multiple cameras (as illustrated in~\cref{fig:1_structure_compare} (c)).
We adopt the standard DETR-like~\cite{DETR3D, Sparse4Dv3} framework, where multi-view images are fed through a shared backbone and the image features are further enhanced by an encoder. 
To effectively identify both 2D and 3D objects, we introduce a hybrid decoder consisting of several multi-view 2D decoder layers and several 3D decoder layers, specifically designed for 2D and 3D object detection tasks, respectively.
3D anchors and queries are used to represent the input of the hybrid decoder.
A Dynamic Query Allocation module assigns 3D anchors to each surrounding image and forms the 2D queries for the 2D decoder layers. The mapping between 3D instances and 2D objects is determined by camera parameters.
After the mapping matrix is established, these 2D queries can be fused by an Adaptive Query Aggregation module to build updated 3D queries for the 3D decoder layers.
To this end, SimPB applies a cyclic 3D-2D-3D approach to interact with both 2D and 3D detection tasks within the hybrid decoder and further boost the overall 2D and 3D detection results.
Moreover, to strengthen the interaction among 2D queries within a particular camera, we introduce a partitioning scheme that groups 2D queries based on the camera index. This partitioning enables us to employ query-group attention operations, which promote effective communication within each camera group and prevent distractions from queries of other cameras.

Our method is evaluated on the nuScenes dataset and archives outstanding results on both 2D and 3D object detection tasks. The contributions of our method can be summarized as:
\begin{itemize}
    \item We propose a novel unified query-based detector that can simultaneously detect 2D objects in the perspective view and 3D objects in the BEV space from multiple cameras. 
    \item We introduce a hybrid decoder for 2D and 3D object detection. 
    A Dynamic Query Allocation and an Adaptive Query Aggregation module are used to continuously update and refine the interaction between 2D and 3D results in a cyclic 3D-2D-3D scheme.  
    \item Our approach achieves promising performance in both multi-view 3D object detection in the BEV space and 2D box detection in image space on the nuScenes dataset.
\end{itemize}

%% file: sec/2_relatedwork.tex
\section{Related Work}

\subsection{2D Object Detection}

2D object detection is a fundamental task in computer vision.
It can be roughly divided into CNN-based methods~\cite{FasterRCNN, CascadeRCNN, LibraRCNN, YoloV1, CenterNet, FCOS} and Transformer-based approaches~\cite{DETR, DeformableDETR, ConditionalDETR, SparseDETR, DAB-DETR, DINO, CO-DETR}. CNN-based methods can be further categorized into two-stage pipeline and one-stage pipeline. Two-stage CNN approaches~\cite{FasterRCNN, CascadeRCNN, LibraRCNN} first create potential regional proposals and then refine them to the accurate bounding box. On the other hand, one-stage methods~\cite{YoloV1, CenterNet, FCOS} directly compute object information and greatly reduce the inference latency. 
More recently, Transformer-based method DETR \cite{DETR} formulates object detection as a set prediction task and eliminates the need for complex post-processing steps. Variants of DETR~\cite{ConditionalDETR, SparseDETR, DAB-DETR, DINO, CO-DETR} have been proposed to enhance training efficiency and further improve object detection performance.
Despite the remarkable achievements in object detection using single input images, the exploration of 2D object detection from multiple cameras remains relatively limited. 
In this paper, we focus on studying multi-view 2D object detection, where the association of 2D objects observed in different cameras is established by utilizing their corresponding 3D information in the BEV space.

\subsection{Multi-view 3D Object Detection}

We group multi-view 3D object detection with surrounding cameras into dense BEV methods~\cite{LSS, BEVDet, BEVDet4D, BEVFormer, BEVFormerV2, BEVDepth, BEVStereo, BEVNeXt, SOLOFusion, VideoBEV, HoP} and sparse query-based methods~\cite{PETR, PETRv2, FocalPETR, StreamPETR, Far3D, DETR4D, Sparse4D, Sparse4Dv2, Sparse4Dv3, SparseBEV, DynamicBEV}.

Dense BEV methods typically construct an explicit BEV feature map from multiple cameras, followed by the 3D object detection sub-model. One example is LSS~\cite{LSS}, which builds intermediate BEV features through interpolation from multi-view image features with a depth estimation step.
The BEVDet series~\cite{BEVDet, BEVDet4D} follows a similar approach to LSS, constructing the BEV space and utilizing additional data augmentation techniques to enhance 3D detection performance.
An alternative method to construct a BEV feature map is through the use of attention layers or MLP operators. For instance, BEVFormer~\cite{BEVFormer, BEVFormerV2} generates the BEV map using learnable grid-shaped BEV queries and employs deformable attention operators. Some dense BEV-based approaches~\cite{BEVDepth, BEVStereo, BEVNeXt} also explore depth supervision for accurate depth estimation to boost object localization.
Furthermore, several works~\cite{SOLOFusion, VideoBEV, HoP} improve object detection performance by propagating long-term historical features and wrapping them to the current timestamp.

Sparse query-based methods directly predict 3D objects from image features using learnable object queries. 
One line of research uses sampled points to establish interactions with multi-view images through camera parameters. 
For example, DETR3D~\cite{DETR3D} adapts the DETR~\cite{DETR} framework by generating a single 3D reference point from an object query using camera parameters to gather image features.
DETR4D~\cite{DETR4D} and the Sparse4D series~\cite{Sparse4D, Sparse4Dv2, Sparse4Dv3} utilize additional reference points from 3D objects to effectively aggregate multi-view image features.
SparseBEV~\cite{SparseBEV} aggregates multi-scale features with adaptive self-attention, allowing for adaptive receptive fields.
DynamicBEV~\cite{DynamicBEV} leverages K-means clustering and top-K Attention to aggregate local and distant features.
Additionally, another approach focuses on learning flexible mappings between queries and image features via attention modules.
PETR~\cite{PETR, PETRv2} employs global attention by incorporating a 3D position encoding that contains geometric information for the mapping.
StreamPETR~\cite{StreamPETR} extends PETR~\cite{PETR} to a long-sequence 3D detection framework by a query propagation mechanism.

\subsection{2D Auxiliary for 3D Object Detection}

Numerous studies have explored the benefits of utilizing 2D detectors as auxiliary information to improve 3D performance. Most approaches perform a detection task on the perspective view and transform the results for 3D query initialization.
For instance, BEVFormer v2~\cite{BEVFormerV2} introduces perspective 3D output as an additional form of supervision. It combines perspective 3D proposals with learnable 3D queries in a two-stage manner to enhance the detection performance.
MV2D~\cite{MV2D} generates 2D detection results on the image view and then lifts these 2D boxes using an implicit sub-network to form 3D queries.
Instead, Far3D~\cite{Far3D} integrates depth estimation with 2D detection results to generate reliable 3D queries. It significantly enhances the perception range and improves the detection performance of distant objects.
Different from them, FocalPETR~\cite{FocalPETR} performs 2D detection on the image view to generate salient foreground feature tokens, which are subsequently used for image feature aggregation along with 3D queries.

Our proposed method, SimPB, differs from the aforementioned approaches in several key aspects. Firstly, while those methods integrate 2D detectors in a two-stage manner, SimPB is an end-to-end and one-stage approach that simultaneously produces both 2D and 3D detection results. 
Secondly, in those approaches, the interaction between 2D and subsequent 3D detectors typically occurs only once in a 2D-3D manner. 
In contrast, SimPB adopts a cyclic 3D-2D-3D approach to interact with both 2D and 3D detectors. This cyclic interaction allows for continuous updates and refinement of the association between 2D and 3D results within the decoder.

%% file: sec/3_method.tex
\begin{figure*}[t]
  \centering
  \includegraphics[width=0.9\linewidth]{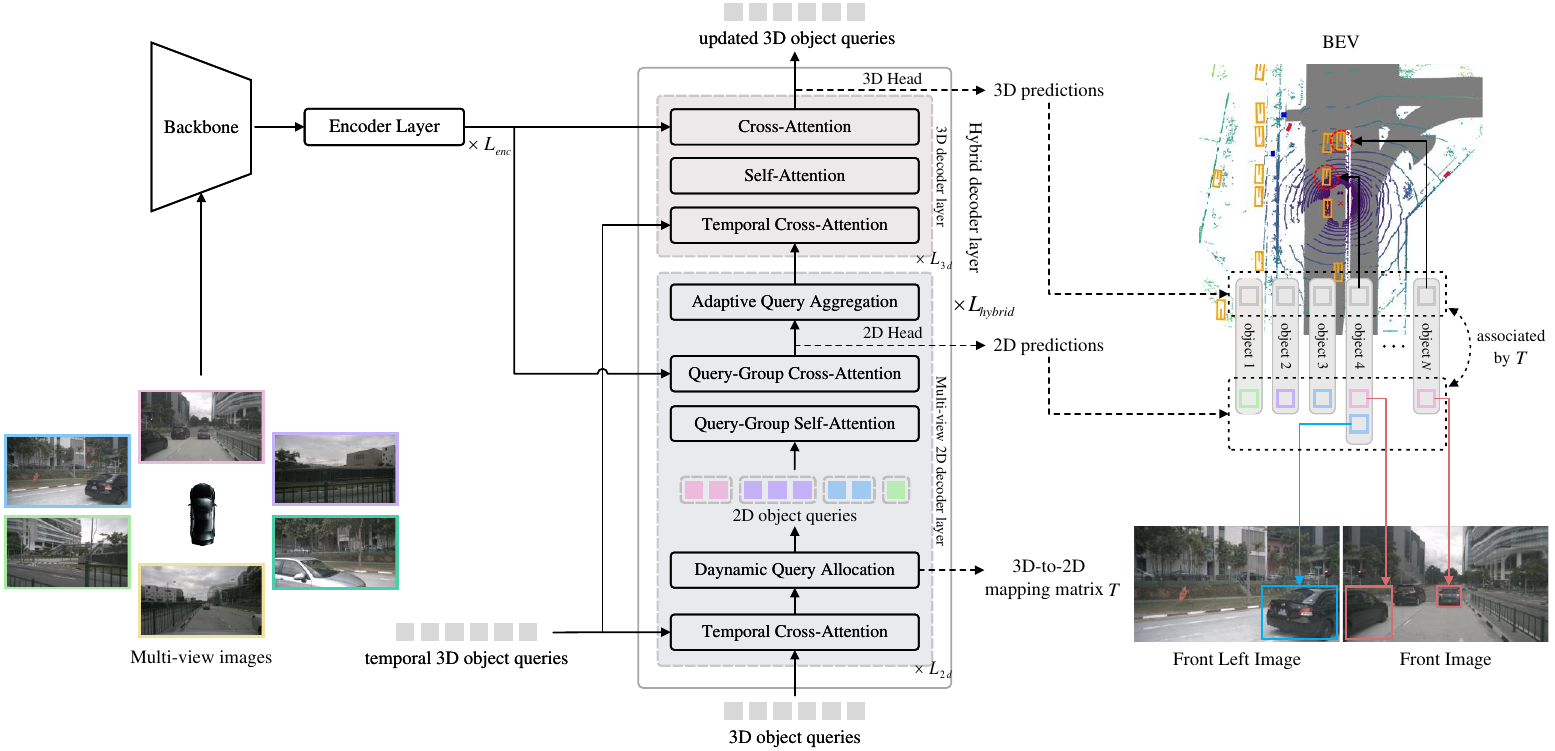}
  \caption{Overview of SimPB, a unified multi-view 2D and 3D object detection framework. 
  Multi-view features are extracted by an image backbone and then enhanced by an encoder module.
  A hybrid decoder module which consists of multi-view 2D decoder layers and 3D decoder layers is used to compute 2D and 3D detection results.}
  \label{fig:2_framework_pipline}
\end{figure*}

\section{Method}

\subsection{Overview}

The overall architecture of our query-based SimPB is shown in ~\cref{fig:2_framework_pipline}. Similar to a DETR-like framework~\cite{DETR, DETR3D, Sparse4Dv3}, SimPB consists of a backbone, an encoder module with $L_{enc}$ layers, and a hybrid decoder module with $L_{hybrid}$ layers.
Given $V$ multi-view images $\{I_i\}_{i=1}^V $, multi-scale features $\{F_i\}_{i=1}^V$ are first extracted by a backbone (e.g. ResNet~\cite{ResNet} or V2-99~\cite{V2-99}). 
We employ an encoder module to enhance multi-scale features, which are then fed into the hybrid decoder module for computing 2D and 3D object results.
Lastly, to fuse temporal information, we follow the method in~\cite{StreamPETR} to propagate top-K history 3D queries frame by frame.

\subsection{Hybrid Decoder Module}

Standard transformer-based 2D detection approaches~\cite{DETR, DeformableDETR, DAB-DETR} apply a 2D decoder layer to extract 2D boxes from the input image. Similarly, popular transformer-based multi-view 3D detection methods~\cite{DETR3D, PETR, StreamPETR, Sparse4Dv3} employ a 3D decoder module to compute 3D objects in the BEV space. Different from them, we propose a hybrid decoder module that simultaneously calculates the 2D objects of each input camera and 3D objects in the BEV space. 
Furthermore, in SimPB, the interaction between 2D and 3D results is continuously updated and refined in a cyclic 3D-2D-3D manner. This cyclic interaction significantly enhances the overall 2D and 3D detection performance across multiple cameras.

Our proposed hybrid decoder module has $L_{2d}$ multi-view 2D decoder layers and $L_{3d}$ 3D decoder layers, each designed to compute 2D boxes in the perspective view and 3D object results in the BEV space, respectively.
The hybrid decoder module takes a set of 3D object queries, denoted as $Q_{3d} \in \mathbb{R}^{N \times C}$, along with their corresponding anchors $A_{3d} = \{(x, y, z, w, l, h, \theta, v_{x}, v_{y})\}$.
Within the multi-view 2D decoder layer, 3D object queries $Q_{3d}$ are passed through a Dynamic Query Allocation block, which allocates them to different camera groups to build 2D object queries $Q_{2d} \in \mathbb{R}^{M \times C}$.
Also, Query-group Self and Cross Attention are utilized to strengthen the interaction among 2D queries within each camera group and compute 2D results for each input camera. 
After obtaining 2D object detection results in the perspective view, these 2D queries are then processed by an Adaptive Query Aggregation block to reconstruct the 3D queries for the subsequent 3D decoder layers. 
The 3D decoder layer utilizes approaches in~\cite{PETR, DynamicBEV, StreamPETR} with self-attention and cross-attention operations and a 3D head to extract 3D objects.
Additionally, we incorporate a temporal cross-attention before both 2D and 3D layers to utilize the historical information by interacting with temporal queries.

\subsection{Dynamic Query Allocation}
\label{sec:allocation}

In the multi-view 2D decoder layer, a Dynamic Query Allocation block is designed to allocate 3D object queries to different camera groups and construct 2D object queries for multi-view 2D detection. 
A simple way is to uniformly distribute $N$ 3D queries to $V$ cameras and generate 2D queries. Instead, we utilize camera parameters to allocate a 3D query to its corresponding cameras and build related 2D queries. 

By interpreting 3D queries as 3D anchors, we can generate a set of $M$ valid 2D queries for $V$ input images. To establish the association between the 3D and 2D queries, we construct a 3D-to-2D mapping matrix $T \in \mathbb{R}^{N \times M}$ based on the camera parameters.
In the mapping matrix $T$, the value $T(i,j) = 1$ indicates that the $i$-th 3D query is associated with the $j$-th 2D query. This matrix enables dynamic allocation and collection of 2D object queries:
\begin{equation}
    Q_{2d} = T^T  \cdot Q_{3d}.
\end{equation}

To determine the validity of a 3D query $q_n$, we begin by projecting its $K$ object points (e.g. center point and eight corner points of a 3D box) to $V$ image planes by camera intrinsic $\mathbf{I} \in \mathbb{R}^{3 \times 3} $ and extrinsic parameters $\mathbf{K} \in \mathbb{R}^{4 \times 4}$.
We obtain a set of points $P_v = \{p_1^v, p_2^v, \ldots, p_K^v \mid p_k^v=(u_k^v, v_k^v)\}$ on $v$-th image view. And $k$ is the index of projected points. 
To determine if the 3D query $q_n$ is valid in the $v$-th image, we define a function $f(q_n, v)$ as:
\begin{equation}
f(q_n,v) = 
\begin{cases} 
    1 & \text{if}~\exists~p_k^v: 0<u_k^v<W, 0<v_k^v<H \\
    0 & \text{otherwise},
\end{cases}
\end{equation}
where $H$ and $W$ are the spatial resolution of an image.

Considering $N$ 3D queries, each camera may have up to $N$ 2D query candidates. We initialize a diagonal matrix $T_C^v \in \mathbb{R}^{N \times N}$ for the $v$-th camera, with the diagonal values set to 1. We retain only the valid 2D queries and eliminate the $j$-th column if $f(q_j, v) = 0$. We construct the mapping $T^v \in \mathbb{R}^{N \times M_v}$ between the 3D queries and the $M_v$ valid 2D queries for the $v$-th camera, derived from $T_C^v$. By concatenating $\{ T^v \}_{v=1}^{V} $ along the column index, we obtain the mapping matrix $T$ between the $N$ 3D queries and the $M$ valid 2D queries.
Totally, we can get $M=\sum_{v=1}^{V}M_v$ valid 2D queries for all the cameras.
In our implementation, the number of $N$ is a predefined parameter of 3D object queries while $M$ is a dynamic number depending on camera parameters and anchor projection.

\begin{figure*}[t]
  \centering
  \includegraphics[width=0.85\linewidth]{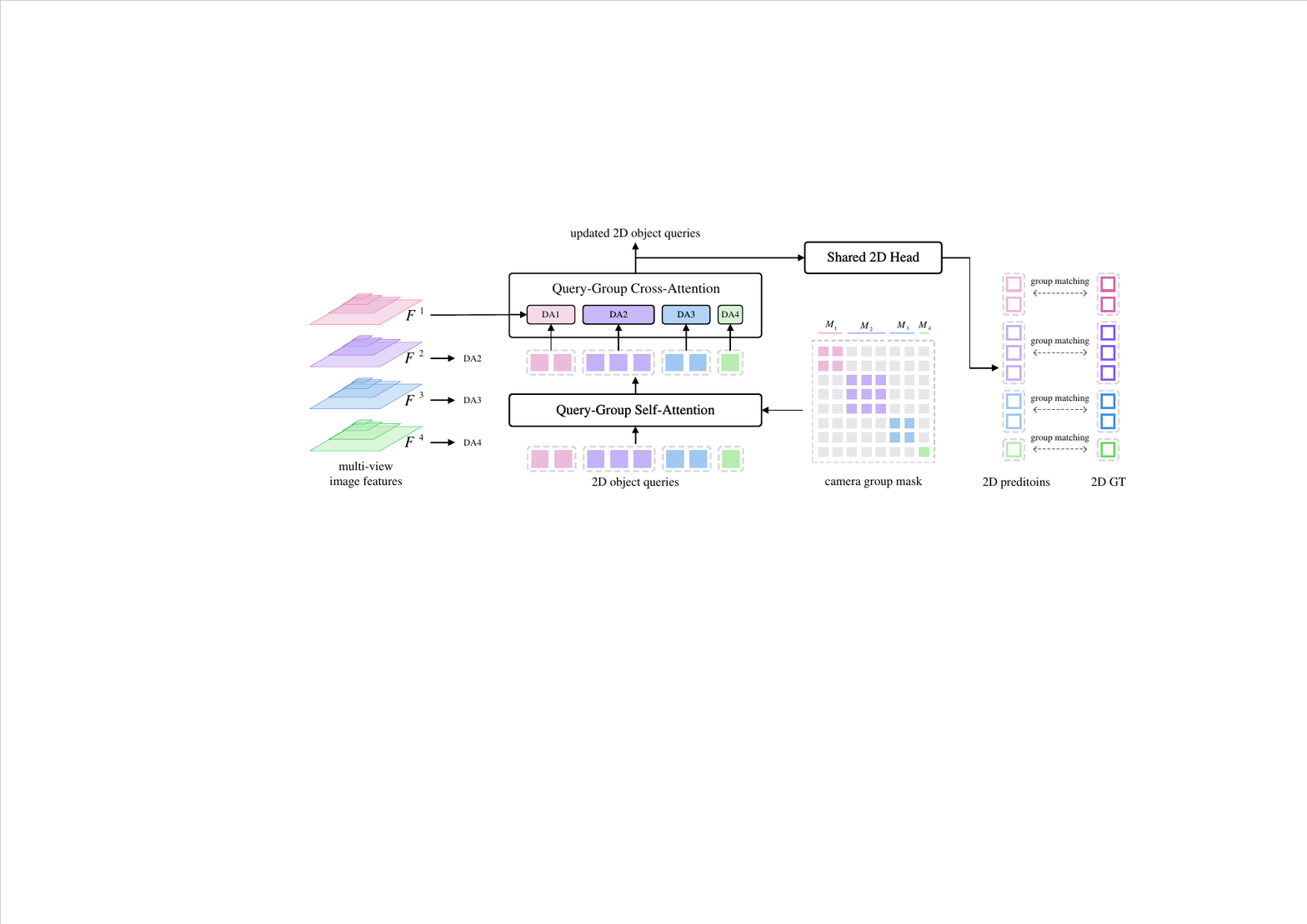}
  \caption{Illustration of the Query-Group Attention.
  We enforce interaction among 2D queries only within the same camera group.
  DA represents deformable attention.}
  \label{fig:4_query_group_attn}
\end{figure*}

\subsection{Query Group Attention}

We adapt DETR-like methods~\cite{DeformableDETR} to extract 2D objects. After dynamic query allocation, we have $M_v$ 2D queries for each camera $v$, resulting in a total of $M$ valid 2D queries.
To strengthen interactions within camera groups, we introduce query-group self-attention and query-group cross-attention.

Specifically, we first create a camera group mask $M_{\text{cg}} \in \mathbb{R}^{M \times M}$ filled with zeros. The value in the mask is then set to 1 if the queries originate from the same camera (as in ~\cref{fig:4_query_group_attn}).
Additionally, we introduce an attention mask $\mathbf{\mathcal{M}} \in \mathbb{R}^{M \times M}$ as:
\begin{align}
\mathbf{\mathcal{M}}(i,j) = 
\left\{
\begin{array}{ll}
  0       & \text{if~} M_{\text{cg}}(i,j)=1\\
  -\infty & \text{otherwise}
\end{array}\right..
\end{align}
This attention mask ensures that queries within the same camera group can attend to each other while preventing queries from different camera groups from attending to one another.

We introduce query-group self-attention, which modifies the standard self-attention mechanism for the 2D query feature $\mathbf{X} \in \mathbb{R}^{M \times C}$ by incorporating an attention mask $\mathbf{\mathcal{M}}$:
\begin{align}
  \mathbf{X} = \text{softmax}(\mathbf{\mathcal{M}} +\frac{\mathbf{Q}\mathbf{K}^{\text{T}}}{\sqrt{C}} )\mathbf{V},
\end{align}
where $\mathbf{Q},~\mathbf{K}~\text{and}~\mathbf{V} \in \mathbb{R}^{M \times C}$ represent the 2D queries $Q_{2d}$ after linear transformations $W_Q$, $W_K$, and $W_V$, respectively. 

Likewise, we can employ a similar method to derive query-group cross-attention. To this end, the proposed group-attention mechanism enables effective information exchange, leading to precise computation of 2D results for each input camera.

In our multi-view 2D decoder, we use the projected object center as the reference point for a 2D box. In the case of object truncation, we rely on the center of the bounding rectangle of the projected anchor. 
Then, we utilize a shared 2D head across cameras to make predictions for the 2D bounding box and class.
Finally, these predictions will be matched with the groundtruth in each groups through standard Hungarian matching algorithm used in DETR-like\cite{DETR, DeformableDETR, DAB-DETR} methods.

\subsection{Adaptive Query Aggregation}
\begin{figure*}[t]
  \centering
  \includegraphics[width=0.88\linewidth]{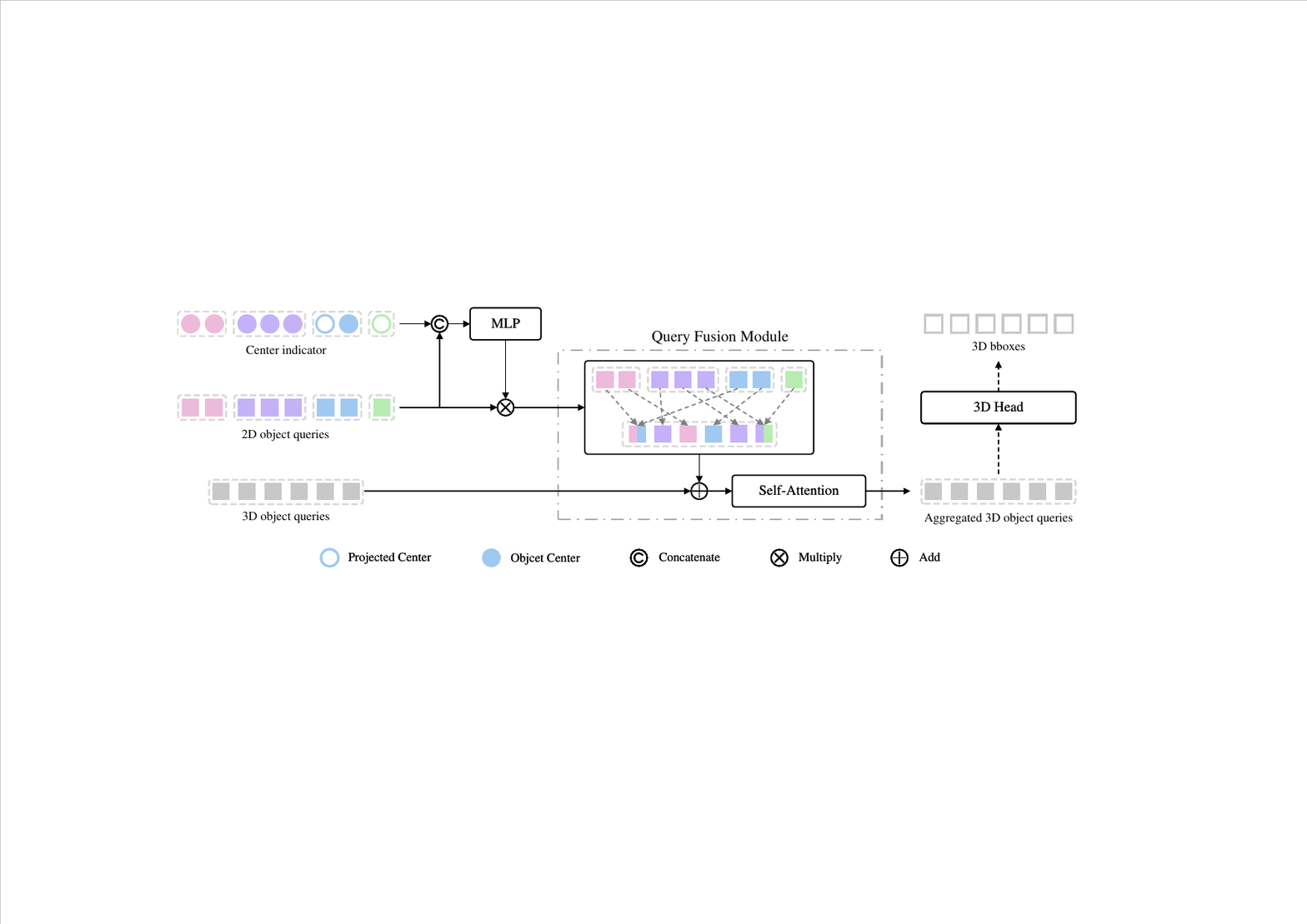}
  \caption{Illusration of the Adaptive Query Aggregation. The indicator vector represents whether a 2D query is truncated or not.}
  \label{fig:5_aggregation_module}
\end{figure*}

After multi-view 2D detection, the 2D object queries are utilized to construct 3D queries for subsequent 3D object detection. As shown in~\cref{fig:5_aggregation_module}, we propose an Adaptive Query Aggregation block to achieve the transformation by using the 3D-to-2D mapping matrix $T$. 

To begin, we extend the 2D queries by incorporating information regarding the truncation status of a 2D object center: $\tilde{Q}_{2d} = Q_{2d} \cdot \text{MLP}(\text{Concat}(Q_{2d}, \mathbbm{1}_{\text{center}}))$, where the indicator $\mathbbm{1}_{\text{center}}$ represents whether a 2D query is truncated or not.

Next, to aggregate $\tilde{Q}_{2d} \in \mathbb{R}^{M \times C}$ into 3D queries, we employ query fusion $Q_{3d}^{\text{agg}} = \phi_{\text{QF}}(\tilde{Q}_{2d}, Q_{3d})$.
Instead of using the simple cross-attention approach, we make use of the 3D-to-2D mapping information:
\begin{equation}
Q_{2d}^{\text{fused}} = \frac{T \cdot \tilde{Q}_{2d}}{\sum_{j=0}^{M}T_j}.
\end{equation}
In this way, the 2D queries, which have been distributed from a shared 3D query during the Dynamic Query Allocation step, are aggregated back together.
Finally, we merge these aggregated queries with the original 3D queries using a residual connection and a self-attention operation: $Q_{3d}^{\text{agg}} = \text{Self-Attn}(Q_{3d} + Q_{2d}^{\text{fused}})$.
Additionally, we employ auxiliary 3D supervision to provide additional guidance for the aggregated 3D object queries.

\subsection{Loss Functions}
The training loss includes both 2D object detection loss and 3D detection loss: $\mathcal{L} = \mathcal{L}_{2d} + \mathcal{L}_{3d}$.
Similar to \cite{Far3D, BEVNeXt, Sparse4Dv3}, our method employ the same 3D detection loss with the depth maps as auxiliary supervision.
For the 2D detection loss, we adopt $\mathcal{L}_{detr2d}$ from DETR-like methods~\cite{DETR, DeformableDETR} and incorporate the observation angle~\cite{alpha_angle} loss of a 3D box in the image view. 
Typically, we encode \textit{alpha}-angle with \textit{sin} and \textit{cos} functions: $\mathcal{L}_{alpha} = \frac{1}{M} \sum_{i=1}^{M} |\sin(\theta_i) - \hat{\sin}(\theta_i)| + |\cos(\theta_i) - \hat{\cos}(\theta_i)|$, where $\theta_i$ is the observation angle of an object. 
And the 2D detection loss is extended to $\mathcal{L}_{2d} = \mathcal{L}_{detr2d} + \lambda_{alpha} \mathcal{L}_{alpha}$, where $\lambda_{alpha}$ is set to 0.5 in the experiments.

%% file: sec/4_experiments.tex
\section{Experiments}
\subsection{Dataset and Metrics}

In the experiments, we evaluate our method using the nuScenes dataset \cite{Nuscene}. This multi-modal videos dataset consists of 1000 videos, which are split into training, validation, and testing sets, consisting of 700, 150, and 150 videos, respectively. Each video is about 20s long, captured by 6 cameras and annotated with Lidar data at a frequency of 2Hz.
The dataset contains a total of 1.4M annotated 3D bounding boxes for 10 classes.
The ground truth of 2D bounding box is generated from 3D labels following methods in~\cite{MV2D, FocalPETR, StreamPETR}.
For 3D object detection, we employ the official evaluation metrics of nuScenes, including mAP, mATE, mASE, mAOE, mAVE, mAAE, and NDS.
Additionally, we evaluate the multi-view 2D object detection using box detection metrics following the standard COCO protocol~\cite{COCODataset} on the nuScenes validation set.

\subsection{Implementation Details}
We follow previous approaches~\cite{Sparse4Dv3} and utilize ResNet50~\cite{ResNet}, ResNet101~\cite{ResNet}, and V2-99~\cite{V2-99} as backbones. To extract multi-view image features, we employ a single deformable transformer encoder layer.
In the hybrid decoder, we set $L_{2d}=1$ and $L_{3d}=1$ in each hybrid layer $L_{hybrid}$ to include one multi-view 2D decoder layer and one 3D decoder layer. The hybrid decoder consists of a total of 6 layers.
We train the models using the AdamW optimizer with an initial learning rate of $4 \times 10^{-4}$. The experiments are conducted with a batch size of 48 on 8 NVIDIA A800 GPUs.
We do not employ Test Time Augmentation (TTA), CBGS\cite{CBGS}, or future frames during training.
The models are trained for 100 epochs in comparison to previously reported methods, while in the ablation study, they are trained for 24 epochs.

\input{table/1_nuscenes_val}

\input{table/2_nuscenes_test}

\input{table/3_nuscenes_val_2D} 

\subsection{Main Results}
\textbf{3D Object Detection Performance.}
We compare SimPB with previous state-of-the-art methods on the validation split of nuScenes in~\cref{tab:val_results3D}.
With a model input resolution of $704\times256$ and ResNet50 as the backbone, SimPB surpasses previous state-of-the-art methods. For example, it outperforms Sparse4Dv3~\cite{Sparse4Dv3} by 0.6\% in mAP and 2.0\% in NDS, also showing significant enhancements of 12.1\% in mAOE.
By using perspective pretrained backbone on nuImages~\cite{Nuscene},
SimPB achieves a new record on the validate split, further improving the performance of DynamicBEV~\cite{DynamicBEV} by 2.3\% in mAP and 2\% in NDS.
SimPB also achieves competitive performance by leveraging a larger ResNet101 backbone and a model input resolution of $1408\times512$. It outperforms Sparse4Dv3 with additional improvements of 0.2\% in mAP and 0.6\% in NDS.

We further evaluate SimPB on the Nuscenes test dataset, as listed in~\cref{tab:tes_results3D}.
With V2-99 as the backbone, SimPB demonstrates a slight improvement in detection performance compared to previous methods such as BEVNext \cite{BEVNeXt} (a dense-BEV based method) and Sparse4Dv3 \cite{Sparse4Dv3} (a sparse query-based method).

It is worth noting that only three 3D decoder layers are specially designed for 3D object detection in SimPB, while the remaining three 2D decoder layers are applied for 2D detection tasks.
In contrast, other approaches typically employ a minimum of six 3D decoder layers for multi-view 3D object detection.

\noindent
\textbf{2D Object Detection Performance.}
\cref{tab:val_results2D} provides a comprehensive comparison of the 2D detection results obtained from multi-view 3D object detection methods that utilize 2D detection results as priors, such as MV2D and StreamPETR~\cite{MV2D, StreamPETR}, as well as the performance of the standard 2D object detector, DeformableDETR~\cite{DeformableDETR}.
The experiments investigate two different backbones and resolutions for the model input. The results demonstrate that SimPB achieves the highest AP score in both model input resolution configurations. Using a smaller model input resolution and ResNet50 as the backbone, 
MV2D provides superior 2D performance for small and medium targets. 
This can be attributed to the inherent strengths of Faster R-CNN~\cite{FasterRCNN}, especially in detecting distant objects in small resolution images.
When using a larger model input and a large backbone, SimPB consistently delivers the best results across all 2D evaluation metrics.
Importantly, our approach distinguishes itself from these methods as it utilizes a single model capable of predicting both 2D and 3D bounding boxes in a unified manner.

\subsection{Ablation Study}

We conduct ablations on the validation split of nuScenes. For all experiments, we use RseNet50 pre-trained on ImageNet~\cite{ImageNet} as the backbone, and the model input resolution is set to $704 \times 256$.

\input{table/4_ablation_structure}

\noindent
\textbf{Choice of Hybrid Decoder.}
The hybrid decoder in our proposed approach is designed with a total number of layers given by $L_{total} = (L_{2d} + L_{3d}) \times L_{hybrid}$. To ensure a fair comparison among different numbers of layers within the hybrid decoder, we limit the total number of layers to 6 by setting $L_{total} = 6$. The results of our ablation study is listed in~\cref{tab:ablation_combination}.
In experiment A, when $L_{2d} = 0$, the hybrid decoder becomes a standard 3D decoder module. In experiment B, when $L_{3d} = 0$, we have only 2D decoder layers with 3D object deep supervision after the Adaptive Query Aggregation block. It shows that, even in these degraded settings, our proposed decoder approach still achieves decent 3D object detection results. 
Interestingly, the model in experiment B, which exclusively utilizes multi-view 2D decoder layers with 3D auxiliary supervision, demonstrates comparable performance to experiment A. This observation emphasizes the effectiveness of our Dynamic Query Allocation and Adaptive Query Aggregation blocks in the proposed multi-view 2D decoder.
By incorporating both 2D decoder and 3D decoder layers (experiments C, D, E, and F), we achieve improved performance compared to the degraded settings.
In experiment E, we also simulate the two-stage scheme of a cascade 2D detector and 3D detector by setting $L_{2d} = 3$ and $L_{3d} = 3$. In the two-stage method, the 2D detector is usually applied to provide initialization of 3D queries. This approach yields superior results compared to model A, which lacks additional 3D query initialization. 
Therefore, the utilization of the multi-view 2D decoder aids in providing more accurate 3D query initialization for the subsequent 3D decoder in the hybrid decoder.
Lastly, model F, which is used in SimPB, employs a cyclic interaction between multi-view 2D and 3D layers and provides the best performance in terms of both mAP and NDS.
In summary, our hybrid decoder successfully exchanges information learned during both 2D and 3D object detection tasks, leading to improved performance of multi-view 3D detection.

\input{table/merge_table_5_6}

\noindent
\textbf{Effect of Dynamic Query Allocation Strategy.} 
In~\cref{tab:ablation_allocation}, we compare different methods of query allocation.
A simple way is to uniformly distribute 3D queries to different cameras without camera information and generate 2D queries. It employs a fixed number of queries for both 2D and 3D objects, without considering the varying number of targets in different cameras. As a result, it overlooks the potential benefits of utilizing 2D objects in the image view. 
We propose three different ways to dynamically allocate 3D anchors to their corresponding cameras using camera parameters. This includes projecting a single object center, an object center with front and rear face centers, and an object center with eight corner points.
These dynamic query allocation methods yield superior results compared to the uniform method. 
In multi-view object detection, objects can appear across different cameras. Therefore, projecting only the single object center to one camera may lead to inaccurate localization of truncated objects in other cameras. To address this limitation, we propose projecting additional points such as the front and rear face centers of an object, as well as the eight corner points of a 3D box. These approaches yield improved results compared to using only a single object center. SimPB which utilizes the object center and eight corners of a 3D box achieves the best results.

\noindent
\textbf{Effect of Adaptive Query Aggregation Strategy.} 
In~\cref{tab:ablation_aggregation}, we explore different methods to fuse 2D queries $\tilde{Q}_{2d}$ and 3D queries $Q_{3d}$. We compare our proposed query fusion approach with a simple cross-attention fusion method. Additionally, we investigate the impact of applying deep supervision to the fused 3D queries $Q_{3d}^{\text{agg}}$ by using 3D ground truth.
The results show that our query fusion approach outperforms the simple cross-attention fusion method. This finding validates the effectiveness of our Adaptive Query Aggregation block in successfully fusing 2D queries to construct 3D queries for the subsequent layer.
Furthermore, incorporating 3D deep supervision for the aggregated 3D queries in the multi-view 2D decoder enhances the detection performance for both methods.

\input{table/merge_table_7_8}

\noindent
\textbf{Impact of observation angle loss.} 
We also investigate the influence of the observation angle loss $\mathcal{L}_{alpha}$. By incorporating $\mathcal{L}_{alpha}$ from the image view into the training loss, we observe a slight improvement in mAP and NDS, along with a significant decrease in mean average orientation error (mAOE). Increasing the weight of $\lambda_{alpha}$ leads to a decrease in mAOE. $\lambda_{alpha}=0.5$ is used in our SimPB.

\noindent
\textbf{Generalization of SimPB.}
In~\cref{tab:ablation_integrate}, we investigate the generalization ability of our method by extending SimPB to other transformer-based approaches, including DETR3D~\cite{DETR3D} and Sparse4Dv2~\cite{Sparse4Dv2}.
We replace certain 3D decoder layers with our multi-view 2D decoder layers in these two methods. 
By incorporating SimPB into DETR3D~\cite{DETR3D}, DETR3D-SimPB achieves an improvement of 2.4\% in mAP and 5.2\% in NDS.
Similarly, Sparse4Dv2-SimPB attains a boost of 3.1\% in mAP and 3.0\% in NDS.
To summarize, SimPB can serve as an extension to existing query-based approaches in multi-view object detection.

%% file: table/1_nuscenes_val.tex
\begin{table*}[t]
\centering
\caption{Comparison results of 3D detection on nuScenes validation dataset. \dag The backbone benefits from perspective pertaining. 
}
\resizebox{0.97\textwidth}{!}  
{
\begin{tabular}{@{}l|c|c|cc|ccccc@{}}
    \toprule
    Method  & \ \, Backbone\ \,  & \ \, Resolution\ \,& \ mAP$\uparrow$\, & \, NDS$\uparrow$\  & \,mATE\,$\downarrow$ & \,mASE\,$\downarrow$ & \,mAOE\,$\downarrow$ & \,mAVE\,$\downarrow$ & \,mAAE\,$\downarrow$ \\
    \midrule
    VideoBEV \cite{VideoBEV}     & ResNet50 &$704\times 256$& 0.422& 0.535& 0.564& 0.276& 0.440& 0.286& 0.198 \\
    SOLOFusion\cite{SOLOFusion}  & ResNet50 &$704\times 256$& 0.427& 0.534& 0.567& 0.274& 0.511& 0.252& \textbf{0.181} \\
    StreamPETR\cite{StreamPETR}  & ResNet50 &$704\times 256$& 0.432& 0.537& 0.609& 0.270& 0.445& 0.279& 0.189 \\
    SparseBEV\cite{SparseBEV}    & ResNet50 &$704\times 256$& 0.432& 0.545& 0.619& 0.283& 0.396& 0.264& 0.194 \\
    BEVNext\cite{BEVNeXt}        & ResNet50 &$704\times 256$& 0.437& 0.548& 0.550& 0.265& 0.427& 0.260& 0.208 \\
    Sparse4Dv2\cite{Sparse4Dv2}  & ResNet50 &$704\times 256$& 0.439& 0.539& 0.598& 0.270& 0.475& 0.282& 0.179 \\
    DynamicBEV\cite{DynamicBEV}  & ResNet50 &$704\times 256$& 0.451& 0.559& 0.606& 0.274& 0.387& 0.251& 0.186 \\
    Sparse4Dv3\cite{Sparse4Dv3}  & ResNet50 &$704\times 256$& 0.469& 0.561& 0.553& 0.274& 0.476& 0.227& 0.200 \\
    \rowcolor{gray!15}
    SimPB\                       & ResNet50 &$704\times 256$& \textbf{0.475}& \textbf{0.581}& \textbf{0.526}& \textbf{0.261}& \textbf{0.355}& \textbf{0.222}& 0.195 \\

    \midrule
    SparseBEV\dag\cite{SparseBEV}   & ResNet50 &$704\times 256$& 0.448& 0.558& 0.595& 0.275& 0.385& 0.253&  \textbf{0.187} \\
    StreamPETR\dag\cite{StreamPETR} & ResNet50 &$704\times 256$& 0.450& 0.550& 0.613& 0.267& 0.413& 0.265& 0.196 \\
    BEVNext\dag\cite{BEVNeXt}       & ResNet50 &$704\times 256$& 0.456& 0.560& \textbf{0.530}& 0.264& 0.424& 0.252& 0.206 \\
    DynamicBEV\dag\cite{DynamicBEV} & ResNet50 &$704\times 256$& 0.464& 0.570& 0.581& 0.271& 0.373& 0.247& 0.190 \\
    \rowcolor{gray!15}
    SimPB\dag   & ResNet50 &$704\times 256$& \textbf{0.487}& \textbf{0.590}& 0.536& \textbf{0.261}& \textbf{0.346}& \textbf{0.208}& \textbf{0.187} \\
    
    \midrule
    SOLOFusion\cite{SOLOFusion}     & ResNet101 &$1408\times 512$& 0.483& 0.582& 0.503& 0.264& 0.381& 0.246& 0.207 \\
    BEVNext\dag\cite{BEVNeXt}       & ResNet101 &$1408\times 512$& 0.500& 0.597& 0.487& 0.260& 0.343& 0.245& 0.197 \\
    SparseBEV\dag\cite{SparseBEV}   & ResNet101 &$1408\times 512$& 0.501& 0.592& 0.562& 0.265& 0.321& 0.243& 0.195 \\
    StreamPETR\dag\cite{StreamPETR} & ResNet101 &$1408\times 512$& 0.504& 0.592& 0.569& 0.262& 0.315& 0.257& 0.199 \\
    Sparse4Dv2\dag\cite{Sparse4Dv2} & ResNet101 &$1408\times 512$& 0.505& 0.594& 0.548& 0.268& 0.348& 0.239& \textbf{0.184} \\
    Far3D\dag\cite{Far3D}           & ResNet101 &$1408\times 512$& 0.510& 0.594& 0.551& 0.258& 0.372& 0.238& 0.195 \\
    DynamicBEV\dag\cite{SparseBEV}  & ResNet101 &$1408\times 512$& 0.512& 0.605& 0.575& 0.270& 0.353& 0.236& 0.198 \\
    
    Sparse4Dv3\dag\cite{Sparse4Dv3} & ResNet101 &$1408\times 512$& 0.537& 0.623& 0.511& \textbf{0.255}& 0.306& 0.194& 0.192 \\
    \rowcolor{gray!15}
    SimPB\dag                       & ResNet101 &$1408\times 512$& \textbf{0.539}& \textbf{0.629}& \textbf{0.475}& 0.260& \textbf{0.280}& \textbf{0.192}& 0.197 \\
    
    \bottomrule
\end{tabular}
}
\label{tab:val_results3D}
\end{table*}

%% file: table/2_nuscenes_test.tex
\begin{table*}[t]
\centering
\caption{Comparison results of 3D detection on nuScenes test dataset.}

\resizebox{0.97\textwidth}{!}  
{
\begin{tabular}{@{}l|c|c|cc|ccccc@{}}
    \toprule
    Method  & \ \, Backbone\ \,  & \ \, Resolution\ \,& \ mAP$\uparrow$\,  & \,NDS$\uparrow$\ & \,mATE\,$\downarrow$ & \,mASE\,$\downarrow$ & \,mAOE\,$\downarrow$ & \,mAVE\,$\downarrow$ & \,mAAE\,$\downarrow$ \\
    \midrule
    StreamPETR\cite{StreamPETR}  & V2-99 &$1600\times 640$& 0.550& 0.636& 0.479& 0.239& 0.317& 0.241& 0.119 \\
    SparseBEV\cite{SparseBEV}    & V2-99 &$1600\times 640$& 0.556& 0.636& 0.485& 0.244& 0.332& 0.246& 0.117 \\
    Sparse4Dv2\cite{Sparse4Dv2}  & V2-99 &$1600\times 640$& 0.556& 0.638& 0.462& 0.238& 0.328& 0.264& 0.115 \\
    BEVNext\cite{BEVNeXt}        & V2-99 &$1600\times 640$& 0.557& 0.642& \textbf{0.409}& 0.241& 0.352& 0.233& 0.129 \\    
    DynamicBEV\cite{DynamicBEV}  & V2-99 &$1600\times 640$& 0.566& 0.648& 0.486& 0.245& 0.343& 0.240& \textbf{0.115} \\  
    Sparse4Dv3\cite{Sparse4Dv3}  & V2-99 &$1600\times 640$& 0.570& 0.656& 0.412& \textbf{0.236}& 0.312& 0.210& 0.117 \\  
    \rowcolor{gray!15}
    SimPB                        & V2-99 &$1600\times 640$& \textbf{0.571}& \textbf{0.660}& 0.419& 0.239& \textbf{0.262}& \textbf{0.201}& 0.133  \\
    \bottomrule
\end{tabular}
}
\label{tab:tes_results3D}
\end{table*}

%% file: table/3_nuscenes_val_2D.tex
\begin{table*}[t]
\centering
\caption{Comparison results of 2D detection on nuScenes val dataset. \dag The backbone benefits from perspective pretraining.}
\resizebox{0.82\textwidth}{!}{  
\begin{tabular}{@{}l|c|c|cccccc@{}}
    \toprule
    Method \ & \, Backbone \, & \, Resolution \, & \, AP \, & \, AP$_{50}$\, & \, AP$_{75}$\, & \, AP$_S$\, & \, AP$_M$ \, & \, AP$_L$ \, \\
    \midrule
    StreamPETR\dag\cite{StreamPETR}         & ResNet50 & $704\times256$ & 0.205 & 0.404 & 0.184 & 0.014 & 0.129 & 0.319 \\
    MV2D\dag\cite{MV2D}                     & ResNet50 & $704\times256$ & 0.226 & 0.456 & 0.198 & \textbf{0.054} & \textbf{0.196} & 0.297 \\
    DeformableDETR\cite{DeformableDETR}     & ResNet50 & $704\times256$ & 0.230 & 0.465 & 0.201 & 0.028 & 0.156 & 0.339 \\
    \rowcolor{gray!15}
    SimPB\dag                               & ResNet50 & $704\times256$ & \textbf{0.256}& \textbf{0.495} & \textbf{0.237} & 0.044 & 0.177 & \textbf{0.361} \\
    \midrule
    StreamPETR\dag\cite{StreamPETR}         & ResNet101 & $1408\times512$ & 0.249 & 0.465 & 0.240 & 0.042 & 0.191 & 0.344 \\
    MV2D\dag\cite{MV2D}                     & ResNet101 & $1408\times512$ & 0.271 & 0.523 & 0.250 & 0.047 & 0.204 & 0.367 \\
    DeformableDETR\cite{DeformableDETR}     & ResNet101 & $1408\times512$ & 0.250 & 0.502 & 0.222 & 0.034 & 0.175 & 0.357 \\
    \rowcolor{gray!15}
    SimPB\dag                               & ResNet101 & $1408\times512$ & \textbf{0.288} & \textbf{0.541} & \textbf{0.276} & \textbf{0.065} & \textbf{0.219} & \textbf{0.388} \\
    \bottomrule
\end{tabular}
}

\label{tab:val_results2D}
\end{table*}

%% file: table/4_ablation_structure.tex
\begin{table*}[t]
\centering
\caption{The ablation studies of different combination of multi-view 2D layer and 3D layer in hybrid decoder layer. }
\resizebox{0.95\textwidth}{!}
{
\begin{tabular}{@{}c|cc|c|cc|ccccc@{}}
    \toprule
    \ Index \ & \ 2D layers \ & \ 3D layers \ & \ Hybrid layers \ &\ mAP$\uparrow$\,  & \,NDS$\uparrow$\ & \,mATE\,$\downarrow$ & \,mASE\,$\downarrow$ & \,mAOE\,$\downarrow$ & \,mAVE\,$\downarrow$ & \,mAAE\,$\downarrow$ \\
    \midrule
    A      & 0 & 1& 6& 0.397 & 0.504 & 0.607 & 0.270 & 0.594 & \textbf{0.270} & 0.196 \\
    B      & 1 & 0& 6& 0.397 & 0.503 & 0.635 & 0.279 & 0.540 & 0.297 & 0.204 \\
    C      & 2 & 1& 2& 0.417 & 0.508 & 0.605 & 0.274 & 0.543 & 0.363 & 0.212 \\
    D      & 1 & 2& 2& 0.419 & 0.517 & 0.599 & \textbf{0.269} & 0.555 & 0.300 & 0.206 \\
    E      & 3 & 3& 1& 0.419 & 0.523 & 0.595 & 0.270 & 0.526 & 0.277 & \textbf{0.192} \\
    \rowcolor{gray!15}
    F      & 1 & 1& 3& \textbf{0.421} & \textbf{0.527} & \textbf{0.590} & 0.274 & \textbf{0.492} & 0.287 & 0.195 \\
    \bottomrule
\end{tabular}
}
\label{tab:ablation_combination}
\end{table*}

%% file: table/merge_table_5_6.tex
\begin{table*}[t]
\begin{floatrow}
\capbtabbox{
\scalebox{0.72}{
  \begin{tabular}{l|cc}
    \toprule
    Allocation Strategy & \ mAP$\uparrow$\,  & \,NDS$\uparrow$\  \\
    \midrule
    Uniform                               & 0.365    & 0.474        \\
    Object Center                         & 0.410    & 0.515        \\
    Object Center + Front-Rear Point      & 0.414    & 0.520        \\
    \rowcolor{gray!15}
    Object Center + Anchor Corner \qquad  & \textbf{0.421}    & \textbf{0.527}        \\
    \bottomrule
\end{tabular}
}}
{
\caption{Ablation studies on query allocation mechanism.}
\label{tab:ablation_allocation}
}
\capbtabbox{
\scalebox{0.72}{
\begin{tabular}{l|cc}
    \toprule
    Aggregation Strategy \ & \ mAP$\uparrow$\,  & \,NDS$\uparrow$\  \\
    \midrule
    Cross-Attention                \  & 0.365     & 0.474    \\
    Cross-Attention + Supervision  \  & 0.399     & 0.495    \\
    Query Fusion                   \  & 0.389     & 0.489    \\
    \rowcolor{gray!15}
    Query Fusion + Supervision     \  & \textbf{0.421}     & \textbf{0.527}    \\
    \bottomrule
\end{tabular}
}
}
{
\caption{Ablation studies on query aggregation mechanism.}
\label{tab:ablation_aggregation}
 \small
}
\end{floatrow}
\end{table*}

%% file: table/merge_table_7_8.tex
\begin{table*}[t]
\begin{floatrow}
\capbtabbox{
\scalebox{0.851}{
\begin{tabular}{@{}l|ccc@{}}
    \toprule
    \ \ $\lambda_{alpha}$   & \ mAP$\uparrow$\,  & \,NDS$\uparrow$ \, & \ mAOE$\downarrow$ \\
    \midrule
    \ \ -            & 0.421          & 0.514 & 0.611  \\
    \ \ 0.25         & \textbf{0.425} & 0.519 & 0.582  \\
    \rowcolor{gray!15}
    \ \ 0.50         & 0.421 & \textbf{0.527} & 0.492  \\
    \ \ 1.00         & 0.418 & 0.523 & \textbf{0.465}  \\
    
    \bottomrule
\end{tabular}
}
}{
\caption{The ablation studies of camera observation loss.}
\label{tab:ablation_loss}
}
\capbtabbox{
\scalebox{0.79}{
\begin{tabular}{l|llc}
    \toprule
    Method \qquad \qquad \qquad  & \ mAP$\uparrow$\,  & \,NDS$\uparrow$\ \\
    \midrule
    DETR3D~\cite{DETR3D}         & \ 0.258   & 0.306   \\
    \rowcolor{gray!15}
    DETR3D-SimPB                 & \ 0.282 (+\hlb{0.024})  & 0.358 (+\hlb{0.052}) \\
    \midrule
    Sparse4Dv2~\cite{Sparse4Dv2} & \ 0.367  & 0.450  \\
    \rowcolor{gray!15}
    Sparse4Dv2-SimPB \           & \ 0.398 (+\hlb{0.031})  & 0.480 (+\hlb{0.030}) \\
    \bottomrule
\end{tabular}
}
}
{
\caption{
Comparison of SimPB with different 3D detectors.
}
\label{tab:ablation_integrate}
 \small
}
\end{floatrow}
\end{table*}

%% file: sec/5_conclusion.tex
\section{Conclusion and Limitation}
In this paper, we introduce a single-stage query-based method for multi-view 2D and 3D object detection. To achieve this, we propose a hybrid decoder module that combines multi-view 2D decoder layers for 2D object detection in the image view and 3D decoder layers for 3D object detection in the BEV space. 
To cyclic interact between 2D and 3D objects in a 3D-2D-3D manner, we propose a dynamic query allocation and adaptive query aggregation module within the hybrid decoder. Additionally, we apply query-group attention to strengthen the interaction among 2D queries within a specific camera.
We extensively evaluate our method, SimPB, on the Nuscenes dataset, conducting comprehensive experiments for both 2D and 3D tasks. 

A limitation of our method is the possibility of increased inference latency due to potential bottlenecks in the allocation process. We will investigate how to improve the inference latency by designing efficient query allocation and aggregation methods.

%% file: sec/supplementary.tex
\appendix
\setcounter{table}{0}   
\setcounter{figure}{0}
\renewcommand{\thetable}{S\arabic{table}}
\renewcommand{\thefigure}{S\arabic{figure}}

\title{SimPB: A Single Model for 2D and 3D Object Detection from Multiple Cameras \\ --------Supplementary Material--------}
\title{Appendix}

\titlerunning{SimPB: A Single Model for 2D and 3D Object Detection}

\author{ }
\authorrunning{Y.~Tang et al.}
\institute{ }

\maketitle
This supplementary material contains additional details of the main manuscript and provides more experiment analysis. 
In~\cref{sec:difference}, we present the difference between SimPB and other previous works that utilize 2D results as priors. 
Next, we elaborate on the complete architecture and give more implementation details in~\cref{sec:implementation}. 
Then, we provide more experiment analysis about runtime, encoder ablation study, and association accuracy in~\cref{sec:experiment}. 
Finally, more visualization results are illustrated in~\ref{sec:qualitative}.

\section{Utilizing 2D Results as Priors}
\label{sec:difference}

\begin{figure}[htpb]
\begin{floatrow}
\ffigbox[\textwidth]{
    \begin{subfloatrow}[3]
    \ffigbox[\FBwidth]{
        \includegraphics[scale=0.42]{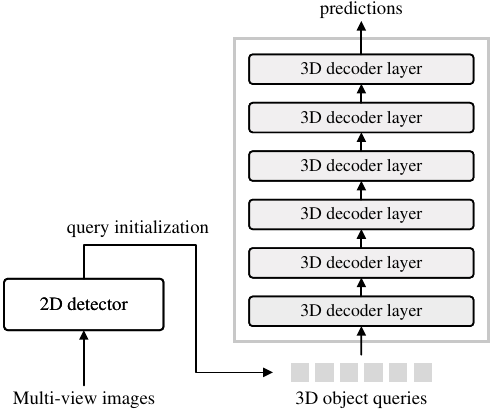}
    }{\caption{query initilization}}
    \ffigbox[\FBwidth]{
        \includegraphics[scale=0.42]{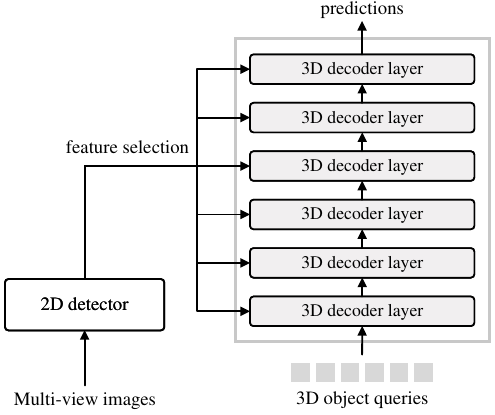}
    }{\caption{feature selection}}
    \ffigbox[\FBwidth]{
    \hspace{10px}
        \includegraphics[scale=0.42]{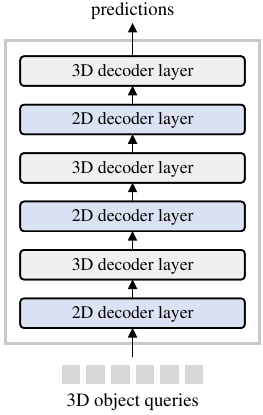}
    \hspace{5px}
    }{\caption{cyclic 2D\&3D layers}}
    \end{subfloatrow}
}{\caption{Comparison of SimPB with previous approaches utilizing 2D results as priors. We roughly categorize these previous methods into two categories, (a) query initialization and (b) feature selection. Instead, SimPB introduces a unified paradigm using novel cyclic 2D\&3D layers as in (c).}
\label{fig:sup_difference_w_previous}}
\end{floatrow}
\end{figure}

We highlight the difference between SimPB and previous approaches that use 2D results as priors in two aspects: architecture and association.

\textbf{Architecture.} 
In \cref{fig:sup_difference_w_previous}, we categorize the previous methods into two groups: query initialization and feature selection, as summarized below.
\begin{itemize}
    \item Query Initialization: In this category, 3D queries are typically initialized from 2D boxes that are detected by a 2D detector (as shown in \cref{fig:sup_difference_w_previous} (a)).
    \item Feature Selection: The methods focus on foreground tokens through 2D supervision and then select them for interaction with 3D queries (as shown in \cref{fig:sup_difference_w_previous} (b)).
\end{itemize}
All these methods employ a 2D detector (or utilize a 2D head) to predict 2D results as a preliminary step before applying a 3D detector.
In contrast, SimPB takes a distinct approach. It performs simultaneous multi-view 2D and 3D detection within a single model using cyclic 2D \& 3D decoder layers (as illustrated in \cref{fig:sup_difference_w_previous} (c)). SimPB is a one-stage method that does not rely on an off-the-shelf 2D detector.

\textbf{Association.} 
For association, we refer to the connection between 2D and 3D results for the same target. A summary of the association of previous methods is listed as follows.
\begin{itemize}
    \item Query Initialization: This method employs a heuristic default association, where a 3D query is linked to a 2D box for its initialization. This association is referred to as a 2D-to-3D association.
    \item Feature Selection: In this approach, the association between 3D queries and selected 2D image tokens, supervised by a 2D detector, is learned through the transformer. However, it does not explicitly establish a direct association between 2D and 3D results.
\end{itemize}
In contrast, our method determines the association by projecting 3D anchors and matching them with the corresponding 2D results. In this way, our approach establishes a 3D-to-2D association between 2D and 3D results. 
The 3D-to-2D association has the advantage of aggregating 2D information more efficiently and avoiding the generation of redundant 3D results. We give a detailed analysis in~\cref{sec:effectiveness_association} and~\cref{sec:qualitative_comparision}.

\section{More Implementation Details}
\label{sec:implementation}
\subsection{Architecture Details}

\begin{figure}[htbp]
\centering
\includegraphics[width=0.95\textwidth]{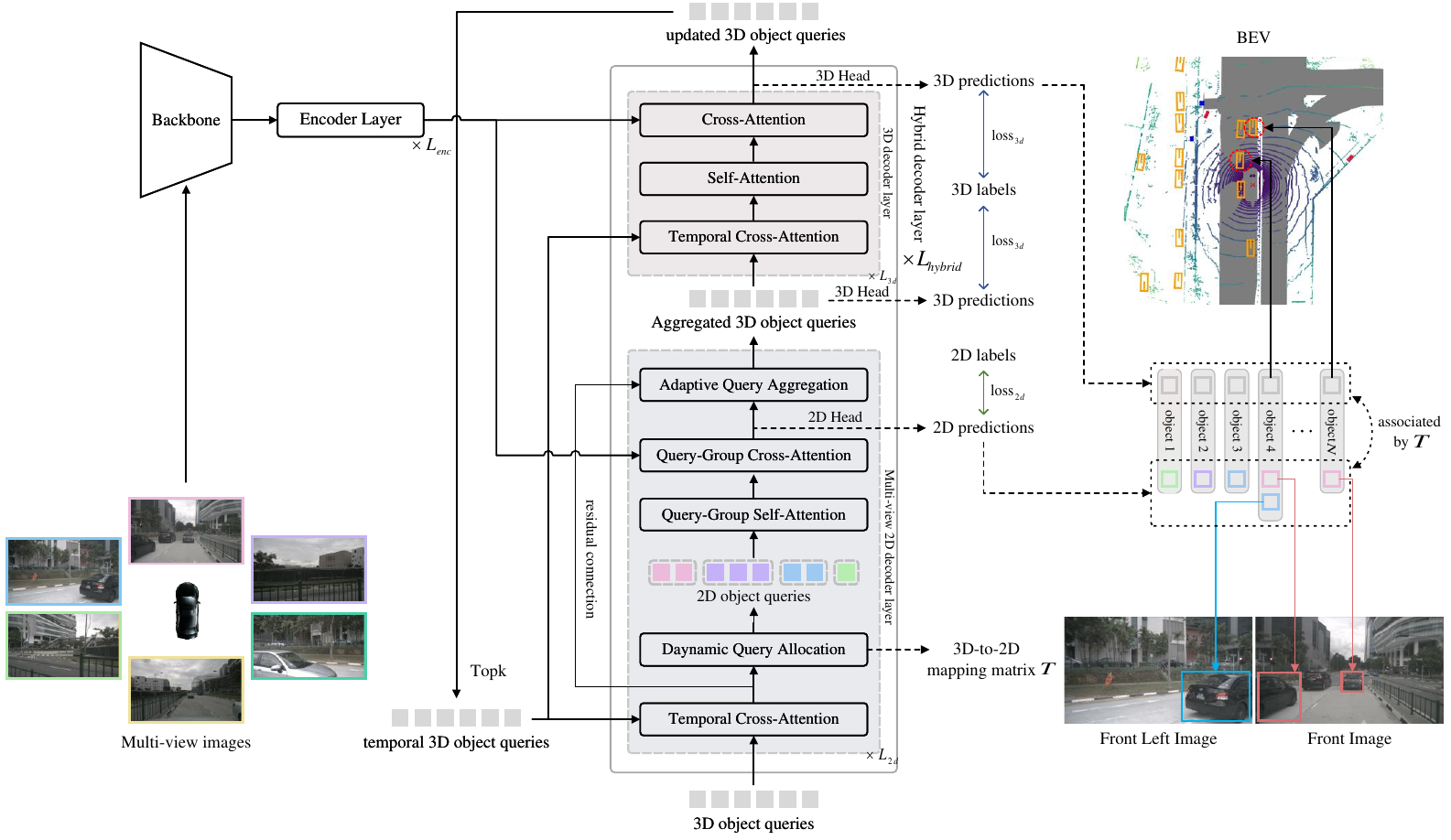}
\caption{Comprehensive architecture of SimPB.}
\label{fig:sup_framework_pipline}
\end{figure}

To maintain clarity, some minor components in Fig. 2 of the main manuscript have been omitted.
We provide the complete architecture of SimPB in \cref{fig:sup_framework_pipline} for a comprehensive view.
Specifically, we include arrow lines to illustrate the connections between self-attention and cross-attention in both the multi-view 2D decoder layer and the 3D decoder layer. 
Additionally, we visualize the residual connection from the output of temporal cross-attention to the Adaptive Query Aggregation module. 
The aggregated 3D queries are separately shown as the output of a multi-view 2D decoder layer, which is used as input for the 3D head for deep supervision. 
Furthermore, to display the temporal propagation, we add an arrow line to indicate the updated object queries linking to the temporal 3D object queries.

\subsection{More Allocation Details}
In the Dynamic Query Allocation module, a 3D query is allocated to a maximum of one object center and multiple projection centers across different camera views by projecting it using camera parameters. The projection center typically represents a truncated portion of a cross-view target. The total number of 2D object queries is equal to the combined count of object centers and projection centers.

During the early stages of training, the presence of inaccurate anchors can lead to a rapid increase in the number of projection centers, resulting in convergence challenges.
To address it, we introduce two constraint strategies to optimize the allocation during training.
\begin{itemize}
    \item The number of projection centers is limited to a maximum of $100$ for each camera group. Consequently, the total number of 2D queries is restricted to a maximum of $N + 100 \times V$, where $N$ represents the number of 3D queries and $V$ denotes the number of cameras.
    
    \item To mitigate the impact of incorrectly projected anchors, we limit the maximum size $\{l, w, h\}$ of the anchors to $\{35, 35, 10\}$, which is computed from the training split of Nuscenes dataset.
\end{itemize}
In our implementation, the number of 3D queries is fixed at $N=900$, while the number of 2D queries $M$ dynamically adjusts based on anchor projection.
During inference, the number of 2D queries $M$ varies around an average of 1100, which is approximately 200 more than the original number of 3D queries $N=900$. 
Nevertheless, this increase in the number of queries introduces only a negligible rise in computational overhead.

\section{More Experimental Analysis}
\label{sec:experiment}

\subsection{Runtime Analysis}
\input{table/sup_runtime_comparision}

We compare the inference speeds of SimPB with state-of-the-art methods using two different backbones and model input resolutions. The evaluation is conducted on an NVIDIA 3090 GPU. 
As shown in~\cref{tab:sup_runtimes_comparision}, SimPB provides inferior performance compared to StreamPETR~\cite{StreamPETR} and Sparse4Dv3~\cite{Sparse4Dv3} at a resolution of $704\times256$. However, at higher resolutions, SimPB achieves comparable inference speeds to these methods. Notably, SimPB consistently outperforms MV2D~\cite{MV2D} in terms of inference speed.

\begin{figure}[htpb]
\begin{floatrow}
\ffigbox[\textwidth]{
    \begin{subfloatrow}[2]
    \ffigbox[\FBwidth]{
        \includegraphics[scale=0.3]{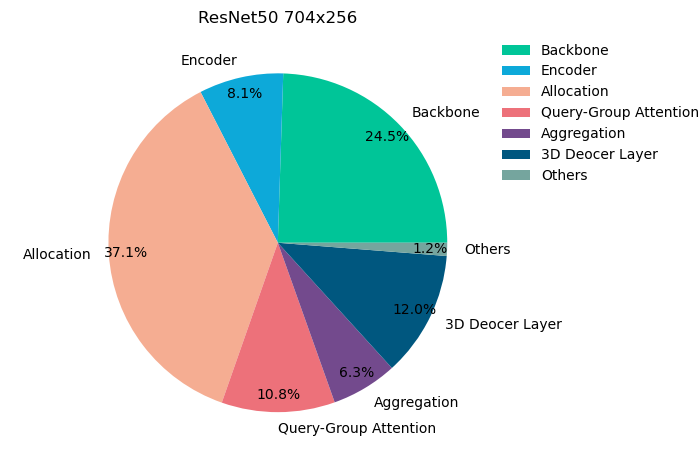}
    }{\caption{ResNet50 \& $704\times256$}}
    \ffigbox[\FBwidth]{
        \includegraphics[scale=0.3]{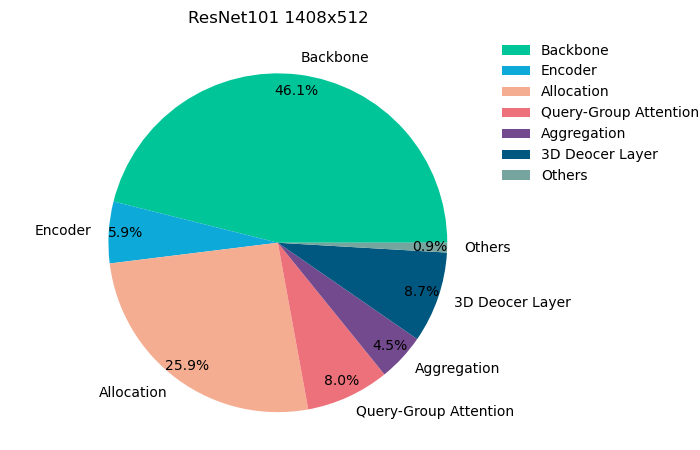}
    }{\caption{ResNet101 \& $1408\times512$}}
        \end{subfloatrow}
}{\caption{Run time decomposition on two configurations.}
\label{fig:sup_runtime_percent}}
\end{floatrow}
\end{figure}

To gain a better understanding of computational complexity, we provide an analysis of the runtime distribution for each module of SimPB under these two settings. The percentage of runtime for each module is illustrated in~\cref{fig:sup_runtime_percent}.
The allocation process in SimPB is responsible for a significant portion of the runtime for the ResNet50 backbone, making it a major bottleneck that affects the overall inference speed.
In the ResNet101 setting, where both the model capacity and the model input increase, the backbone itself takes up more time and becomes a significant bottleneck. However, the processing time for the allocation step does not vary with changes in model size or input.
As a result, the model experiences a relatively smaller negative impact when utilizing a larger backbone and higher resolution. We plan to optimize the inference speed of the network in our future work.

\subsection{Impact of Encoder}
\input{table/sup_ablation_encoder}
Our objective is to develop a unified architecture capable of simultaneously output both
2D and 3D results from multiple cameras. To achieve this, we adopt an encoder-decoder structure following the design of the DETR-like scheme~\cite{DeformableDETR}.
In contrast to previous sparse query-based methods such as DETR3D~\cite{DETR3D}, PETR~\cite{PETR}, and Sparse4D~\cite{Sparse4D}, which often exclude the encoder for 3D detection, we recognize the potential benefits of incorporating an encoder. 

To explore its effectiveness, we conduct an additional ablation study, as listed in~\cref{tab:sup_ablation_encoder}. The experimental settings align with Sec. 4.4 of the main manuscript.
By incorporating the encoder, the model boosts the performance in mAP, NDS, and $\text{AP}_{2d}$.
Increasing the number of encoder layers further enhances performance but at the cost of increased computation time and memory consumption.
To achieve a balance between efficiency and effectiveness, we employ a single encoder layer in SimPB. 

\subsection{Effect of Association}
\label{sec:effectiveness_association}

\begin{figure}[htpb]
\begin{floatrow}
\ffigbox[\textwidth]{
    \begin{subfloatrow}[2]
    \ffigbox[\FBwidth]{
        \includegraphics[scale=0.37]{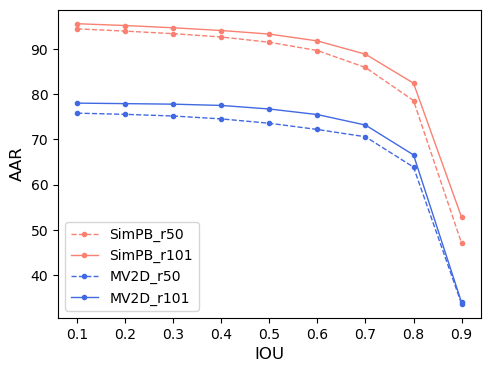}
    }{\caption*{}\label{subfig:a}}
    \ffigbox[\FBwidth]{
        \includegraphics[scale=0.37]{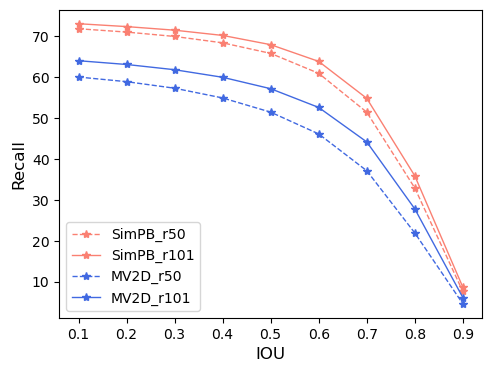}
    }{\caption*{}\label{subfig:b}}
    \end{subfloatrow}
}{\caption{AAR (Association Accuracy Rate) and Recall curves.}
\label{fig:sup_association_metric}}
\end{floatrow}
\end{figure}

Two-stage methods usually only provide a default association between 3D and 2D results during query initialization. In contrast, SimPB explicitly constructs the association of 2D and 3D detection results. To quantitatively evaluate the association, we design a metric termed Association Accuracy Rate (AAR) to measure the accuracy rate of association and also apply Recall as an evaluation metric as well. 

Suppose there are $N_{\text{3d}}$ 3D grountruth boxes $\{G_{3d}^{i}\}_{i=1}^{N_{\text{3d}}}$ and $N_{\text{2d}}$ projected 2D boxes $\{G_{2d}^{i}\}_{i=1}^{N_{\text{2d}}}$ on the image views as 2D boxes label in the validation dataset. We can obtain $M_{\text{3d}}$ 3D box prediction $\{P_{3d}^{i}\}_{i=1}^{M_{\text{3d}}}$ and $M_{\text{2d}}$ 2D prediction $\{P_{2d}^{i}\}_{i=1}^{M_{\text{2d}}}$ from the network.
Typically, a candidate match is established between a 3D prediction and a 2D groundtruth box. The total number of candidates matching is denoted as $\# \text{Matching}_\text{\{3D-pred, 2D-gt\}}$. From these candidate matches, we select valid associations between 3D predictions and 2D predictions generated by the network. The valid matching number is $\# \text{ValidMatching}_\text{\{3D-pred, 2D-pred\}}$. Therefore, we define the association evaluation metric AAR as follows:
\begin{gather}
    \text{AAR} =\cfrac{\# \text{ValidMatching}_\text{\{3D-pred, 2D-pred\}}}{\# \text{Matching}_\text{\{3D-pred, 2D-gt\}}} \times 100\%, \\
    \text{Recall} = \cfrac{\# \text{Matching}_\text{\{3D-pred, 2D-gt\}}}{N_{2d}} \times 100\%.
\end{gather} 

For a given 3D prediction $P_{3d}^{i}$ and $j$-th 2D groundtruth $G_{2d}^{j}$ on $v$-th image view. And we denote the associated 3D grountruth of $G_{2d}^{j}$ as $G_{3d}^{j}$ for simplicity. 
$P_{3d \rightarrow 2d}^{i}$ is the bounding rectangle on $v$-th view projected from $P_{3d}^{i}$.
The connection between $P_{3d}^{i}$ and $G_{2d}^{j}$ is a candidate matching if the following conditions are met:
\begin{equation}
\begin{scriptsize}
\Phi(P_{3d}^{i}, G_{2d}^{j}) =
\begin{cases} 
    1 & \text{if}~\text{Dist}(P_{3d}^{i} , G_{3d}^{j} ) \leq \tau_{dis} 
    ~\&~ \text{IoU}(P_{3d \rightarrow 2d}^{i} , G_{2d}^{j} ) \geq \tau_{iou}  
    ~\&~ \text{Cls}(P_{3d}^{i}, G_{3d}^{j})=1 \\
    0 & \text{otherwise},
\end{cases}
\end{scriptsize}
\end{equation}
where $\text{Dist}$ represents the Euclidean distance between the centers of two 3D boxes, and $\text{Cls}$ indicates whether the labels of the two boxes are the same or not.
Similarly, we denote a valid candidate between the $i$-th 3D prediction $P_{3d}^{i}$ and the $k$-th 2D prediction $P_{2d}^{k}$ when the following conditions are met:
\begin{equation}
\begin{scriptsize}
\Psi(P_{3d}^{i}, P_{2d}^{k}) =
\begin{cases} 
    1 & \text{if}~\Phi(P_{3d}^{i}, G_{2d}^{j})=1
    ~\&~ \text{IoU}(P_{2d}^{k} , G_{2d}^{j} ) \geq \tau_{iou}  
    ~\&~ \text{Cls}(P_{2d}^{k}, G_{2d}^{j})=1 \\
    0 & \text{otherwise}.
\end{cases}
\end{scriptsize}
\end{equation}
Therefore, AAR can be rewritten as:
\begin{equation}
    \text{AAR}  =\cfrac{\sum_{i=1}^{M_{3d}}\sum_{k=1}^{M_{2d}} \Psi(P_{3d}^{i}, P_{2d}^{k}) } {\sum_{i=1}^{M_{3d}}\sum_{j=1}^{N_{2d}} \Phi(P_{3d}^{i}, G_{2d}^{j}) } \times 100\%  
\end{equation}

We fix the $\tau_{dis} = 2$ and adjust $\tau_{iou}$ from 0.1 to 0.9 to draw the AAR and Recall curves.
As shown in \cref{fig:sup_association_metric}, the accuracy and recall decrease as the IOU threshold $\tau_{iou}$ increases.
However, SimPB constantly gains higher AAR and Recall metrics with a large margin on both settings.
The utilization of the 3D-to-2D association in SimPB demonstrates higher accuracy compared to the 2D-to-3D association used in MV2D. This approach not only maintains a larger number of matched predictions but also ensures better alignment of 2D and 3D features. As a result, the 2D information to the same target is effectively leveraged for 3D tasks, leading to improved performance.

\section{Qualitative Evaluation}
\label{sec:qualitative}

\begin{figure}[!htp]
\ffigbox[\textwidth]
{
\begin{subfloatrow}[1]
\ffigbox[\FBwidth]{
    \includegraphics[width=0.9\textwidth]{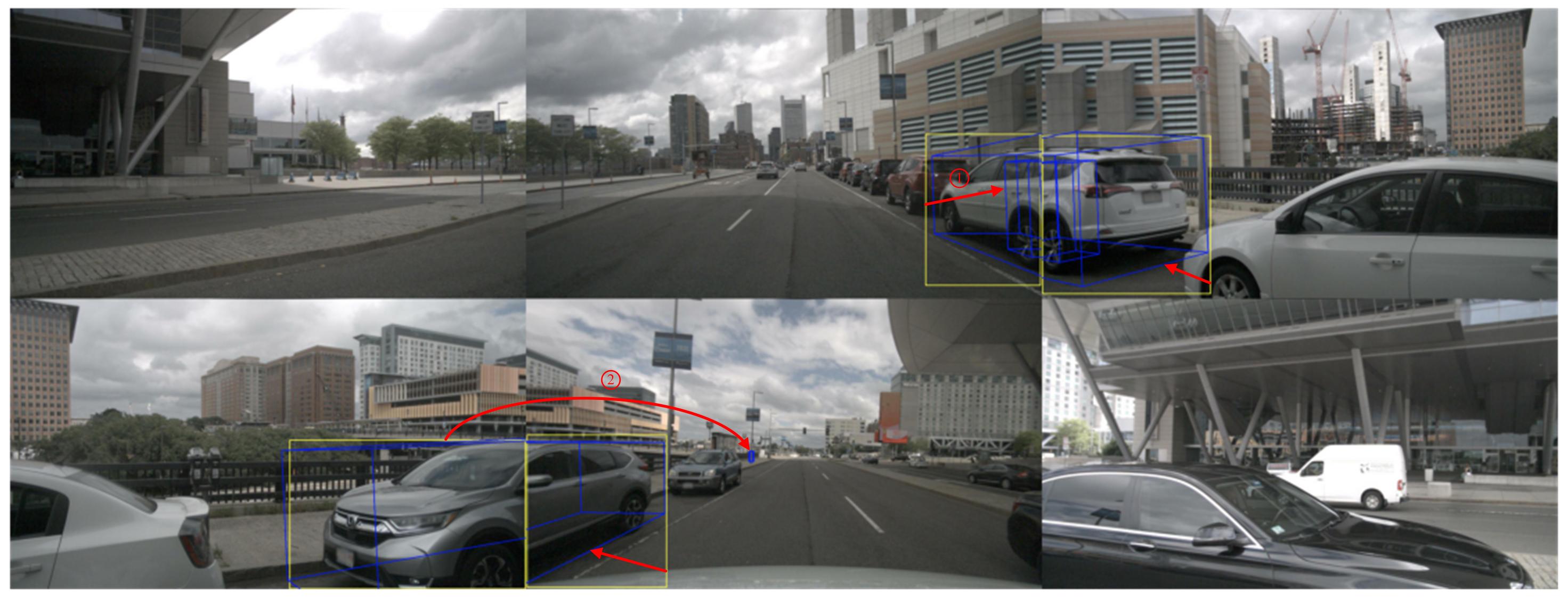}
}
{\caption{The detection results of MV2D. 2D-to-3D association (\hlr{red arrow}) may produce duplicate 3D results or unrelated results from 2D priors for a cross-camera target.
}}
\end{subfloatrow}    

\begin{subfloatrow}[1] 
\ffigbox[\FBwidth]{
\vspace{10px}
    \includegraphics[width=0.9\textwidth]{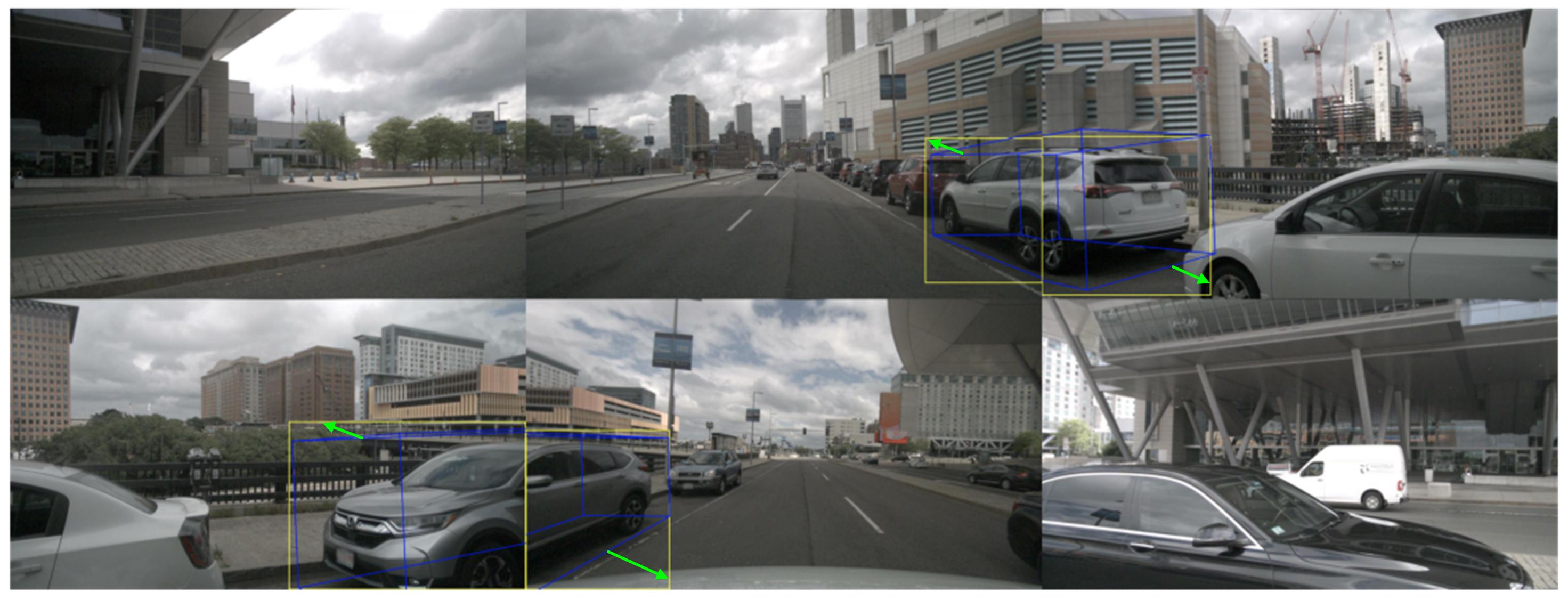}
}{\caption{The detection results of SimPB. 
The process of 3D-to-2D association (\hlg{green arrow}) effectively yields accurate 3D results along with their corresponding 2D boxes for cross-camera targets.
}}
\end{subfloatrow}
}{
\caption{Illustration of association between 2D and 3D results by MV2D and SimPB.}
\label{fig:sup_association_comparision}
}
\end{figure} 

\subsection{Qualitative Comparison on Association}
\label{sec:qualitative_comparision}

To conduct a qualitative analysis of the association establishment between 2D-to-3D and 3D-to-2D, we compare the detection results of MV2D and SimPB in the same keyframe. 
The detection results of several cross-camera targets are shown in~\cref{fig:sup_association_comparision}, where the yellow boxes represent the 2D detection results, and the blue boxes represent the 3D detection results

In the case of a cross-camera target $O$, MV2D initially employs a 2D detector to generate multiple 2D bounding boxes (using two as an example). 
These 2D results are used to initialize 3D queries through a 2D-to-3D association method. However, multiple 3D queries are associated with the target $O$. Only one of these 3D queries accurately predicts the target, while the other may produce a duplicated nearby object (circle 1 in~\cref{fig:sup_association_comparision}~(a)) or even an unrelated result (circle 2 in~\cref{fig:sup_association_comparision}~(a)).
This discrepancy arises during the Hungarian matching step, where only the best candidate query is optimized as the positive sample, resulting in the suppression of the remaining 3D queries. Consequently, the 2D information from a specific view of the suppressed 3D query is discarded, despite it can provide relevant information about the same target.

To address this issue, SimPB adopts a novel approach to establish the association between 2D and 3D results using a 3D-to-2D method. For a cross-camera target, we distribute its 3D queries to different views for 2D detection tasks and subsequently aggregate the results to form a single 3D query.
To this end, for a cross-camera object, SimPB only produces one 3D detection result along with its corresponding 2D detection in each relevant camera. Also, our cyclic 3D-2D-3D interaction ensures there is a single coherent representation of the target across different views, eliminating redundancy outputs and enhancing the accuracy of the results (in~\cref{fig:sup_association_comparision}~(b)).

\subsection{Qualitative Comparison with State-of-the-Art Methods}

\begin{figure}[htbp]
\centering
\includegraphics[width=\textwidth]{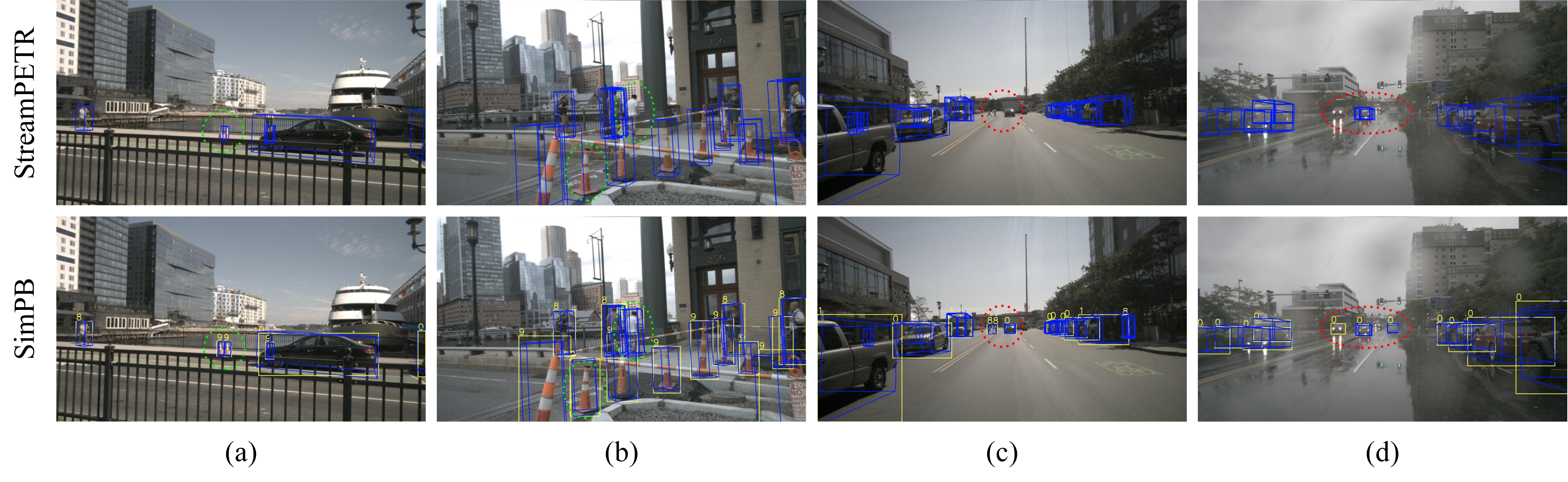}
\caption{Visualization results of StreamPETR and SimPB.}
\label{fig:sup_compare_with_streampetr}
\end{figure}

SimPB provides improved accuracy in detecting crowd objects such as traffic cones and pedestrians compared to StreamPETR. For instance, while StreamPETR incorrectly identifies two traffic cones as a single entity, SimPB accurately detects them as separate objects (green circle in~\cref{fig:sup_compare_with_streampetr}~(a)).
In Figure~\ref{fig:sup_compare_with_streampetr}~(b), StreamPETR also provides an inaccurate estimation of the locations of pedestrians and traffic cones, showing that crowd objects tend to cluster around their neighboring objects. In contrast, SimPB provides more precise results and successfully distinguishes crowded and small objects. This improvement can be attributed to the novel cyclic 3D-2D-3D scheme of SimPB, where the iterative and interactive process of 2D and 3D information enhances the refinement of queries, resulting in more accurate detection results.

SimPB also demonstrates its advantage in detecting distant targets and performs well even in challenging scenarios. 
For example, SimPB successfully detects pedestrians and cars at far distances, whereas StreamPETR fails to do so (red circle in~\cref{fig:sup_compare_with_streampetr}~(c)).
Furthermore, despite encountering difficulties in predicting small and distant targets within complex environments, such as rain (as shown in~\cref{fig:sup_compare_with_streampetr}~(d)), SimPB can still provide reliable 2D detections. These 2D detections can be utilized in subsequent post-processing steps within a practical autonomous driving perception system.

\subsection{More Visualization Results}
\begin{figure}[!htp]
\ffigbox[\textwidth]
{
\begin{subfloatrow}[1]
\ffigbox[\FBwidth]{
    \includegraphics[width=0.95\textwidth]{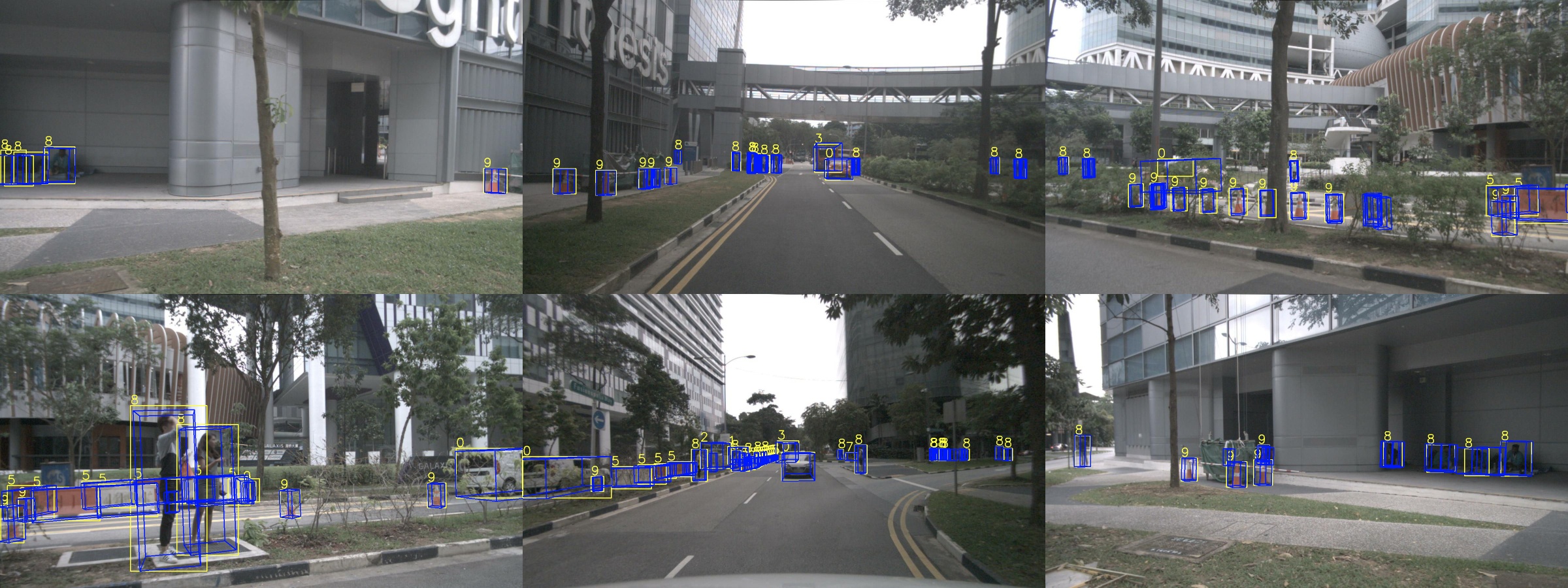}
}{\caption*{}}
\vspace{-15px}
\end{subfloatrow}    

\begin{subfloatrow}[1] 
\ffigbox[\FBwidth]{
    \includegraphics[width=0.95\textwidth]{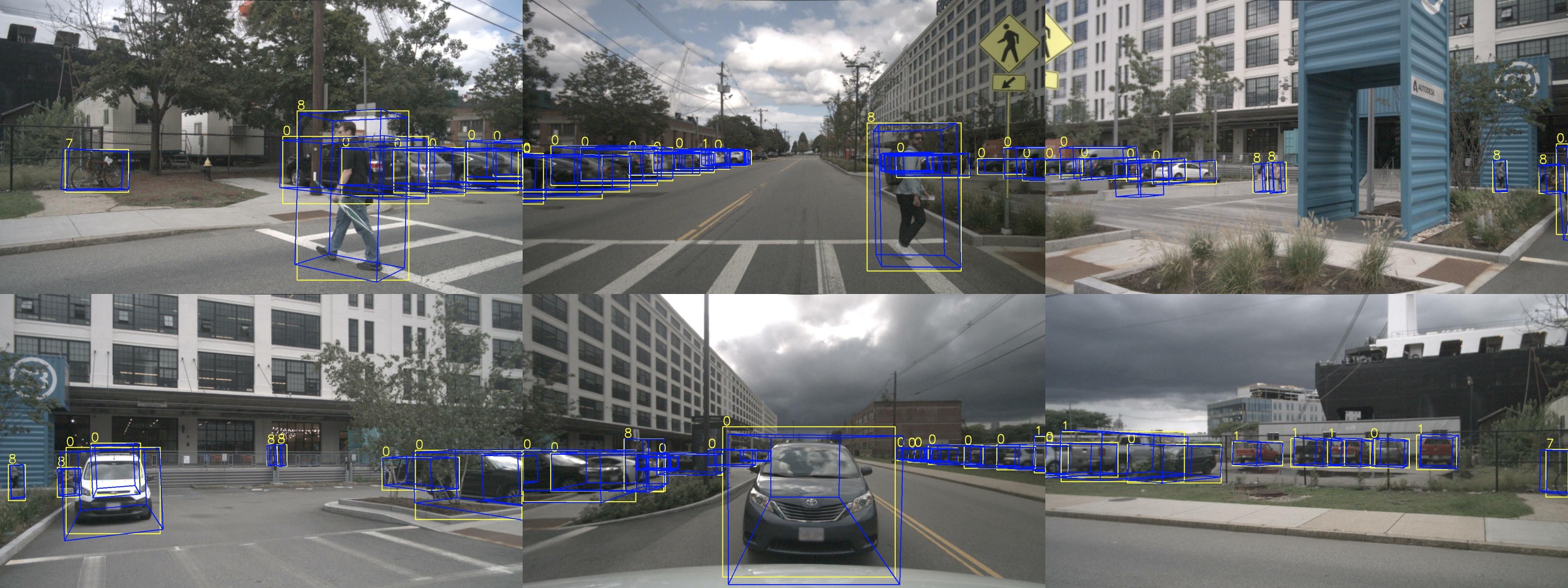}
}{\caption*{}}
\vspace{-15px}

\end{subfloatrow}
\begin{subfloatrow}[1] 
\ffigbox[\FBwidth]{
    \includegraphics[width=0.95\textwidth]{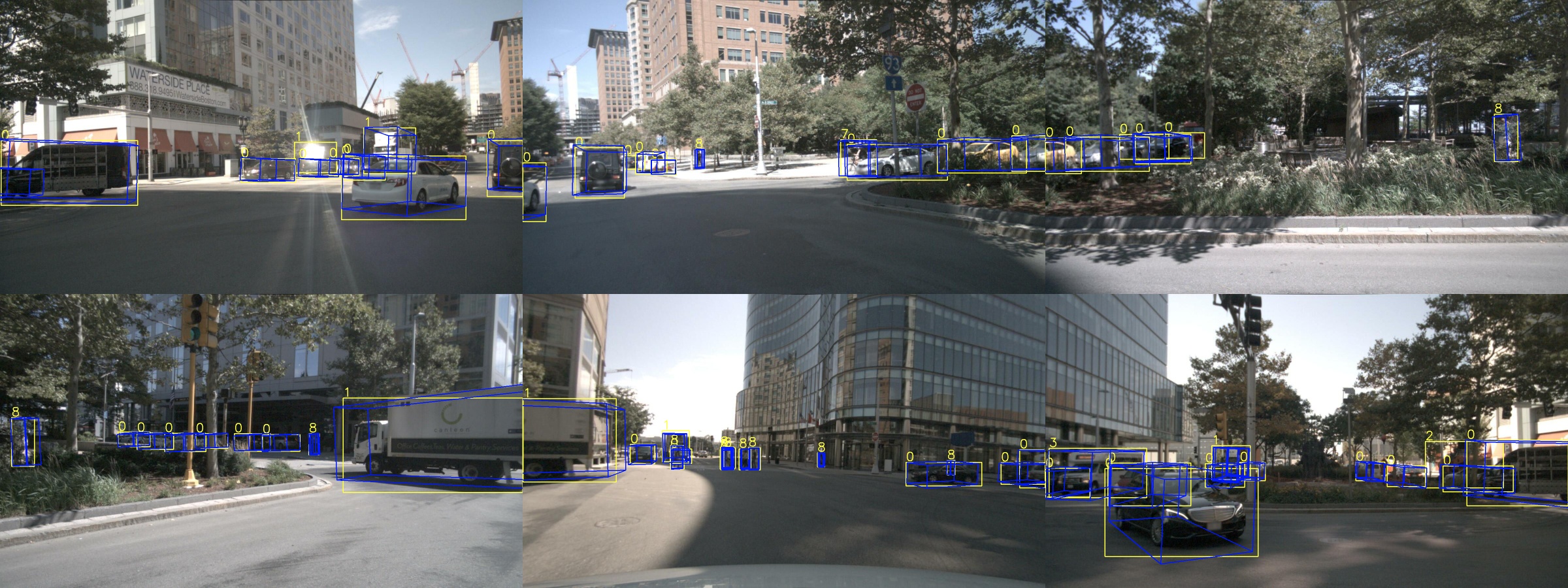}
}{\caption*{}}
\end{subfloatrow}
}{
\caption{Detection results on Nuscenes validation dataset during the daytime. 2D predict results are visualized in \hly{yellow} and 3D results are visualized in \hlb{bule}.}
\label{fig:sup_visualization1}
}
\end{figure}

\begin{figure}[!htp]
\ffigbox[\textwidth]
{
\begin{subfloatrow}[1]
\ffigbox[\FBwidth]{
    \includegraphics[width=0.95\textwidth]{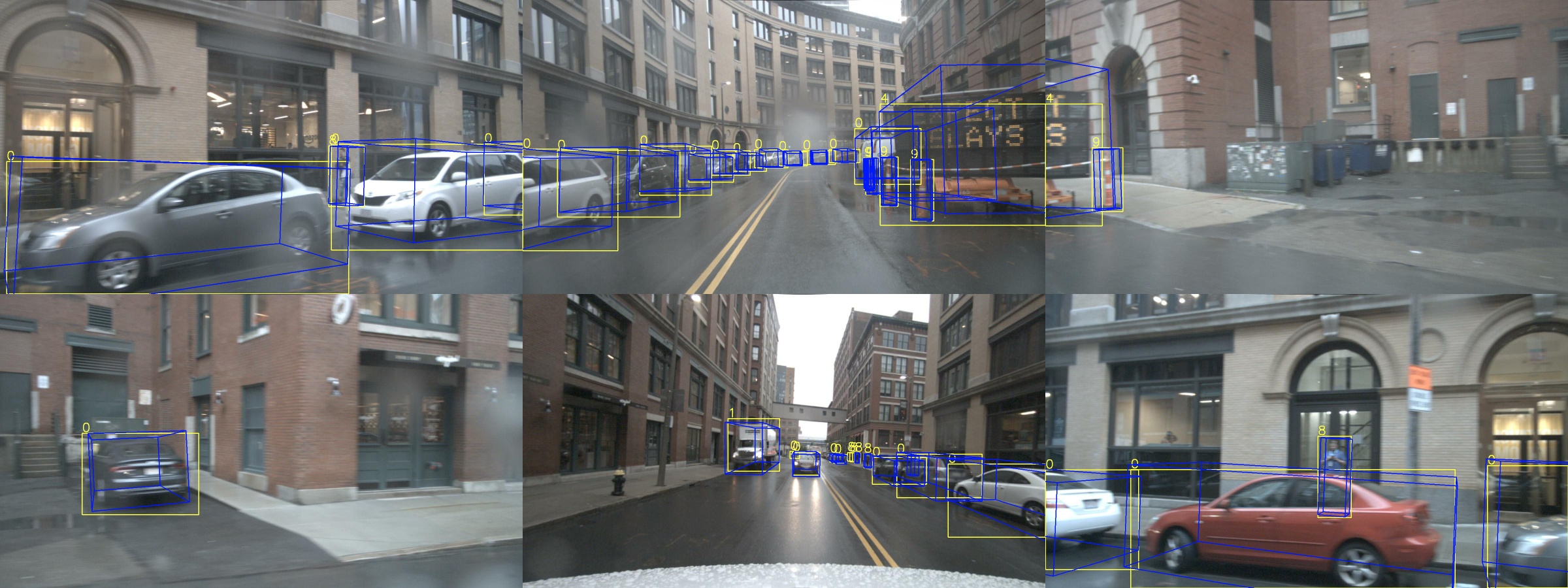}
}{\caption*{}}
\vspace{-15px}
\end{subfloatrow}    

\begin{subfloatrow}[1] 
\ffigbox[\FBwidth]{
    \includegraphics[width=0.95\textwidth]{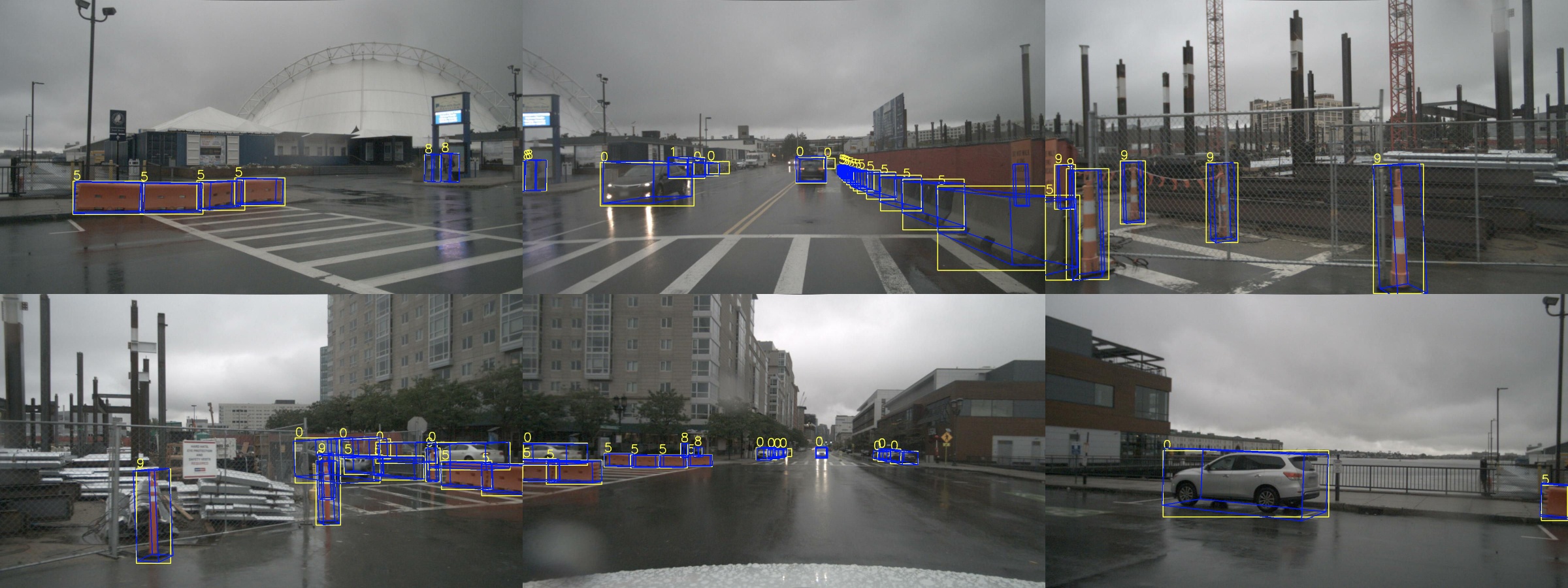}
}{\caption*{}}
\vspace{-15px}

\end{subfloatrow}
\begin{subfloatrow}[1] 
\ffigbox[\FBwidth]{
    \includegraphics[width=0.95\textwidth]{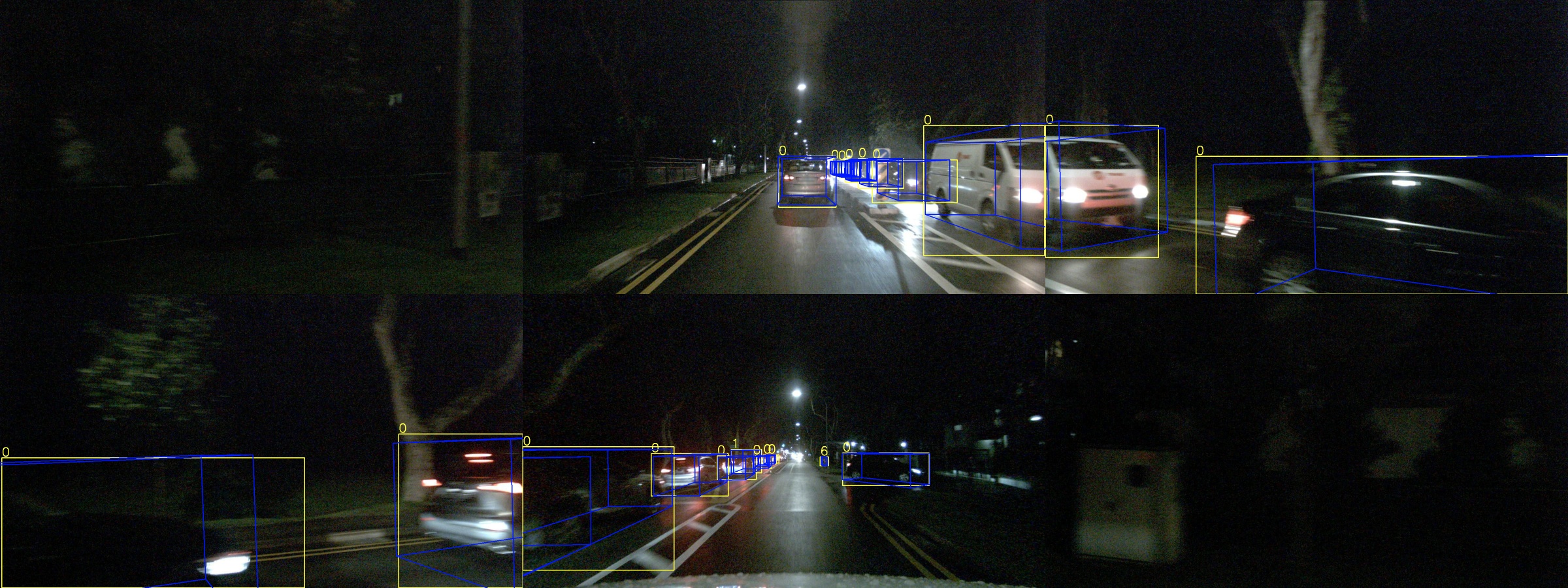}
}{\caption*{}}
\end{subfloatrow}
}{
\caption{Detection results on Nuscenes validation dataset during the rain and night. 2D predict results are visualized in \hly{yellow} and 3D results are visualized in \hlb{bule}.}
\label{fig:sup_visualization2}}
\end{figure}

We present the visualization of the 2D and 3D detection results of SimPB using the ResNet101 backbone and a model input resolution of $1408 \times 512$. The visualizations are shown in~\cref{fig:sup_visualization1} and~\cref{fig:sup_visualization2}. The number on the detected box represents its predicted category.

%% file: table/sup_runtime_comparision.tex
\begin{table}[htbp]
\centering
\caption{
Comparison of inference speeds.
}
\scalebox{0.85}{
\begin{tabular}{@{}l|cc|c@{}}
    \toprule
    \ Method \  & \, Backbone\,  & \, Resolution \, & \, FPS$\uparrow$ \,\\
    \midrule
    \ MV2D\cite{MV2D}             \  & ResNet50  & $704 \times 256$ & 9.5   \\
    \ StreamPETR\cite{StreamPETR} \  & ResNet50  & $704 \times 256$ & 27.1   \\
    \ Sparse3Dv3\cite{Sparse4Dv3} \  & ResNet50  & $704 \times 256$ & 19.8   \\
    \rowcolor{gray!15}
    \ SimPB \                        & ResNet50  & $704 \times 256$ & 10.9   \\
    \midrule
    \ MV2D\cite{MV2D}             \  & ResNet101  & $1408 \times 512$ & 3.9   \\
    \ StreamPETR\cite{StreamPETR} \  & ResNet101  & $1408 \times 512$ & 6.4   \\
    \ Sparse3Dv3\cite{Sparse4Dv3} \  & ResNet101  & $1408 \times 512$ & 8.2   \\
    \rowcolor{gray!15}
    \ SimPB                       \  & ResNet101  & $1408 \times 512$ & 7.1   \\
    \bottomrule
\end{tabular}
}
\label{tab:sup_runtimes_comparision}
\end{table}

%% file: table/sup_ablation_encoder.tex
\begin{table}[htbp]
\centering
\caption{
Impact of encoder layer of SimPB.
}
\scalebox{0.85}{
\begin{tabular}{@{}c|ccccc@{}}
    \toprule
    \ Encoder layers \  & \ mAP$\uparrow$\,  & \,NDS$\uparrow$ \, & \,$\text{AP}_{2d}\uparrow$ \,&  FPS$\uparrow$ \ & \,Memory(G)\,$\downarrow$ \\
    \midrule
    \ - \        & 0.412 & 0.519 & 0.211 & 12.9 &  10.64\\
    \rowcolor{gray!15}
    \ 1 \        & 0.421 & 0.527 & 0.217 & 10.9 &  14.46\\
    \ 2 \        & 0.425 & 0.529 & 0.220 & 9.0  &  17.89\\
    \ 3 \        & 0.432 & 0.536 & 0.222 & 7.9  &  21.07\\ 
    \ 4 \        & 0.439 & 0.545 & 0.224 & 6.1  &  23.94\\ 
    \bottomrule
\end{tabular}
}
\label{tab:sup_ablation_encoder}
\end{table}

%% file: main.bbl
\begin{thebibliography}{10}
\providecommand{\url}[1]{\texttt{#1}}
\providecommand{\urlprefix}{URL }
\providecommand{\doi}[1]{https://doi.org/#1}

\bibitem{M3D-RPN}
Brazil, G., Liu, X.: M3d-rpn: Monocular 3d region proposal network for object
  detection. In: ICCV. pp. 9287--9296 (2019)

\bibitem{Nuscene}
Caesar, H., Bankiti, V., Lang, A.H., Vora, S., Liong, V.E., Xu, Q., Krishnan,
  A., Pan, Y., Baldan, G., Beijbom, O.: nuscenes: A multimodal dataset for
  autonomous driving. In: CVPR. pp. 11618--11628 (2020)

\bibitem{CascadeRCNN}
Cai, Z., Vasconcelos, N.: Cascade r-cnn: Delving into high quality object
  detection. In: CVPR. pp. 6154--6162 (2018)

\bibitem{Monocular3DOD}
Chen, X., Kundu, K., Zhang, Z., Ma, H., Fidler, S., Urtasun, R.: Monocular 3d
  object detection for autonomous driving. In: CVPR. pp. 2147--2156 (2016)

\bibitem{ImageNet}
Deng, J., Dong, W., Socher, R., Li, L.J., Li, K., Fei-Fei, L.: Imagenet: A
  large-scale hierarchical image database. In: CVPR. pp. 248--255. Ieee (2009)

\bibitem{CenterNet}
Duan, K., Bai, S., Xie, L., Qi, H., Huang, Q., Tian, Q.: Centernet: Keypoint
  triplets for object detection. In: ICCV. pp. 6569--6578 (2019)

\bibitem{VideoBEV}
Han, C., Sun, J., Ge, Z., Yang, J., Dong, R., Zhou, H., Mao, W., Peng, Y.,
  Zhang, X.: Exploring recurrent long-term temporal fusion for multi-view 3d
  perception. arXiv preprint arXiv:2303.05970  (2023)

\bibitem{ResNet}
He, K., Zhang, X., Ren, S., Sun, J.: Deep residual learning for image
  recognition. In: Proceedings of the IEEE conference on computer vision and
  pattern recognition. pp. 770--778 (2016)

\bibitem{BEVDet4D}
Huang, J., Huang, G.: Bevdet4d: Exploit temporal cues in multi-camera 3d object
  detection. arXiv preprint arXiv:2203.17054  (2022)

\bibitem{BEVDet}
Huang, J., Huang, G., Zhu, Z., Ye, Y., Du, D.: Bevdet: High-performance
  multi-camera 3d object detection in bird-eye-view. arXiv preprint
  arXiv:2112.11790  (2021)

\bibitem{Far3D}
Jiang, X., Li, S., Liu, Y., Wang, S., Jia, F., Wang, T., Han, L., Zhang, X.:
  Far3d: Expanding the horizon for surround-view 3d object detection. In: AAAI
  (2023)

\bibitem{V2-99}
Lee, Y., Hwang, J.w., Lee, S., Bae, Y., Park, J.: An energy and gpu-computation
  efficient backbone network for real-time object detection. In: CVPR (2019)

\bibitem{BEVStereo}
Li, Y., Bao, H., Ge, Z., Yang, J., Sun, J., Li, Z.: Bevstereo: Enhancing depth
  estimation in multi-view 3d object detection with temporal stereo. In: AAAI.
  pp. 1486--1494 (2023)

\bibitem{BEVDepth}
Li, Y., Ge, Z., Yu, G., Yang, J., Wang, Z., Shi, Y., Sun, J., Li, Z.: Bevdepth:
  Acquisition of reliable depth for multi-view 3d object detection. In: AAAI.
  pp. 1477--1485 (2023)

\bibitem{BEVNeXt}
Li, Z., Lan, S., Alvarez, J.M., Wu, Z.: Bevnext: Reviving dense bev frameworks
  for 3d object detection. In: CVPR (2024)

\bibitem{BEVFormer}
Li, Z., Wang, W., Li, H., Xie, E., Sima, C., Lu, T., Qiao, Y., Dai, J.:
  Bevformer: Learning bird’s-eye-view representation from multi-camera images
  via spatiotemporal transformers. In: ECCV. pp. 1--18 (2022)

\bibitem{COCODataset}
Lin, T.Y., Maire, M., Belongie, S., Hays, J., Perona, P., Ramanan, D.,
  Doll{\'a}r, P., Zitnick, C.L.: Microsoft coco: Common objects in context. In:
  ECCV. pp. 740--755. Springer (2014)

\bibitem{Sparse4D}
Lin, X., Lin, T., Pei, Z., Huang, L., Su, Z.: Sparse4d: Multi-view 3d object
  detection with sparse spatial-temporal fusion. arXiv preprint
  arXiv:2211.10581  (2022)

\bibitem{Sparse4Dv2}
Lin, X., Lin, T., Pei, Z., Huang, L., Su, Z.: Sparse4d v2: Recurrent temporal
  fusion with sparse model. arXiv preprint arXiv:2305.14018  (2023)

\bibitem{Sparse4Dv3}
Lin, X., Pei, Z., Lin, T., Huang, L., Su, Z.: Sparse4d v3: Advancing end-to-end
  3d detection and tracking. arXiv preprint arXiv:2311.11722  (2023)

\bibitem{SparseBEV}
Liu, H., Teng, Y., Lu, T., Wang, H., Wang, L.: Sparsebev: High-performance
  sparse 3d object detection from multi-camera videos. In: ICCV. pp.
  18580--18590 (2023)

\bibitem{DAB-DETR}
Liu, S., Li, F., Zhang, H., Yang, X., Qi, X., Su, H., Zhu, J., Zhang, L.:
  Dab-detr: Dynamic anchor boxes are better queries for detr. In: ICLR (2022)

\bibitem{PETR}
Liu, Y., Wang, T., Zhang, X., Sun, J.: Petr: Position embedding transformation
  for multi-view 3d object detection. In: ECCV. pp. 531--548 (2022)

\bibitem{PETRv2}
Liu, Y., Yan, J., Jia, F., Li, S., Gao, Q., Wang, T., Zhang, X., Sun, J.:
  Petrv2: A unified framework for 3d perception from multi-camera images. arXiv
  preprint arXiv:2206.01256  (2022)

\bibitem{DETR4D}
Luo, Z., Zhou, C., Zhang, G., Lu, S.: Detr4d: Direct multi-view 3d object
  detection with sparse attention. arXiv preprint arXiv:2212.07849  (2022)

\bibitem{ConditionalDETR}
Meng, D., Chen, X., Fan, Z., Zeng, G., Li, H., Yuan, Y., Sun, L., Wang, J.:
  Conditional detr for fast training convergence. In: ICCV. pp. 3651--3660
  (2021)

\bibitem{alpha_angle}
Mousavian, A., Anguelov, D., Flynn, J., Kosecka, J.: 3d bounding box estimation
  using deep learning and geometry. In: Proceedings of the IEEE conference on
  Computer Vision and Pattern Recognition. pp. 7074--7082 (2017)

\bibitem{DETR}
Nicolas, C., Francisco, M., Gabriel, S., Nicolas, U., Alexander, K., Sergey,
  Z.: End-to-end object detection with transformers. In: ECCV. pp. 213–--229
  (2020)

\bibitem{LibraRCNN}
Pang, J., Chen, K., Shi, J., Feng, H., Ouyang, W., Lin, D.: Libra r-cnn:
  Towards balanced learning for object detection. In: CVPR. pp. 821--830 (2019)

\bibitem{SOLOFusion}
Park, J., Xu, C., Yang, S., Keutzer, K., Kitani, K., Tomizuka, M., Zhan, W.:
  Time will tell: New outlooks and a baseline for temporal multi-view 3d object
  detection. In: ICLR (2023)

\bibitem{LSS}
Philion, J., Fidler, S.: Lift, splat, shoot: Encoding images from arbitrary
  camera rigs by implicitly unprojecting to 3d. In: ECCV. pp. 194--210 (2020)

\bibitem{YoloV1}
Redmon, J., Divvala, S., Girshick, R., Farhadi, A.: You only look once:
  Unified, real-time object detection. In: CVPR. pp. 779--788 (2016)

\bibitem{FasterRCNN}
Ren, S., He, K., Girshick, R., Sun, J.: Faster r-cnn: Towards real-time object
  detection with region proposal networks. IEEE TPAMI pp. 1137--1149 (2017)

\bibitem{SparseDETR}
Roh, B., Shin, J., Shin, W., Kim, S.: Sparse detr: Efficient end-to-end object
  detection with learnable sparsity. In: ICLR (2021)

\bibitem{FCOS}
Tian, Z., Shen, C., Chen, H., He, T.: Fcos: Fully convolutional one-stage
  object detection. In: ICCV. pp. 9627--9636 (2019)

\bibitem{FocalPETR}
Wang, S., Jiang, X., Li, Y.: Focal-petr: Embracing foreground for efficient
  multi-camera 3d object detection. IEEE Transactions on Intelligent Vehicles
  (2023)

\bibitem{StreamPETR}
Wang, S., Liu, Y., Wang, T., Li, Y., Zhang, X.: Exploring object-centric
  temporal modeling for efficient multi-view 3d object detection. In: ICCV. pp.
  3621--3631 (2023)

\bibitem{FCOS3D}
Wang, T., Zhu, X., Pang, J., Lin, D.: Fcos3d: Fully convolutional one-stage
  monocular 3d object detection. In: ICCV. pp. 913--922 (2021)

\bibitem{DETR3D}
Wang, Y., Guizilini, V., Zhang, T., Wang, Y., Zhao, H., Solomon, J.: Detr3d: 3d
  object detection from multi-view images via 3d-to-2d queries. In: COLR. pp.
  180--191 (2021)

\bibitem{MV2D}
Wang, Z., Huang, Z., Fu, J., Wang, N., Liu, S.: Object as query: Equipping any
  2d object detector with 3d detection ability. In: ICCV. pp. 3791--3800 (2023)

\bibitem{BEVFormerV2}
Yang, C., Chen, Y., Tian, H., Tao, C., Zhu, X., Zhang, Z., Huang, G., Li, H.,
  Qiao, Y., Lu, L., Zhou, J., Dai, J.: Bevformer v2: Adapting modern image
  backbones to bird’s-eye-view recognition via perspective supervision. In:
  CVPR. pp. 17830--17839 (2023)

\bibitem{DynamicBEV}
Yao, J., Lai, Y.: Dynamicbev: Leveraging dynamic queries and temporal context
  for 3d object detection. arXiv preprint arXiv:2310.05989  (2023)

\bibitem{DINO}
Zhang, H., Li, F., Liu, S., Zhang, L., Su, H., Zhu, J., Ni, L.M., Shum, H.Y.:
  Dino: Detr with improved denoising anchor boxes for end-to-end object
  detection. In: ICLR (2022)

\bibitem{CBGS}
Zhu, B., Jiang, Z., Zhou, X., Li, Z., Yu, G.: Class-balanced grouping and
  sampling for point cloud 3d object detection. arXiv preprint arXiv:1908.09492
   (2019)

\bibitem{DeformableDETR}
Zhu, X., Su, W., Lu, L., Li, B., Wang, X., Dai, J.: Deformable detr: Deformable
  transformers for end-to-end object detection. In: ICLR (2020)

\bibitem{HoP}
Zong, Z., Jiang, D., Song, G., Xue, Z., Su, J., Li, H., Liu, Y.: Temporal
  enhanced training of multi-view 3d object detector via historical object
  prediction. In: ICCV (2023)

\bibitem{CO-DETR}
Zong, Z., Song, G., Liu, Y.: Detrs with collaborative hybrid assignments
  training. In: ICCV. pp. 6748--6758 (2023)

\end{thebibliography}
